\documentclass{article} %
\usepackage[final]{colm2025_conference}

\usepackage{microtype}
\usepackage{hyperref}
\usepackage{url}
\usepackage{booktabs}

\usepackage{amsmath}
\usepackage{multirow}
\usepackage{graphicx}
\usepackage{cleveref}
\usepackage{lineno}
\usepackage{colortbl}
\usepackage{tabularx}

\definecolor{darkblue}{rgb}{0, 0, 0.5}
\hypersetup{colorlinks=true, citecolor=darkblue, linkcolor=darkblue, urlcolor=darkblue}

\title{Rethinking Multilingual Continual Pretraining: Data Mixing for Adapting LLMs Across Languages and Resources}

\author{Zihao Li$^{1}$, Shaoxiong Ji$^{1,2,3}$\thanks{Corresponding author. Work done while at the University of Helsinki.}, Hengyu Luo$^{1}$, J\"org Tiedemann$^{1}$  \\
        $^1$University of Helsinki \quad $^2$ELLIS Institute Finland \quad $^3$ University of Turku \\
\texttt{\{zihao.li,~shaoxiong.ji,~hengyu.luo,~jorg.tiedemann\}@helsinki.fi}
}

\begin{document}

\maketitle

\begin{abstract}
Large Language Models (LLMs) exhibit significant disparities in performance across languages, primarily benefiting high-resource languages while marginalizing underrepresented ones. Continual Pretraining (CPT) has emerged as a promising approach to address this imbalance, although the relative effectiveness of monolingual, bilingual, and code-augmented data strategies remains unclear. This study systematically evaluates 36 CPT configurations involving three multilingual base models, across 30+ languages categorized as altruistic, selfish, and stagnant, spanning various resource levels. Our findings reveal three major insights: (1) Bilingual CPT improves multilingual classification but often causes language mixing issues during generation. (2) Including programming code data during CPT consistently enhances multilingual classification accuracy and language modeling capabilities, particularly benefiting low-resource languages, but introduces a trade-off by slightly degrading generation quality. (3) Contrary to prior work, we observe substantial deviations from language classifications according to their impact on cross-lingual transfer: Languages classified as altruistic often negatively affect related languages, selfish languages show conditional and configuration-dependent behavior, and stagnant languages demonstrate surprising adaptability under certain CPT conditions. These nuanced interactions emphasize the complexity of multilingual representation learning, underscoring the importance of systematic studies on generalizable language classification to inform future multilingual CPT strategies.
\end{abstract}

\section{Introduction}
Large Language Models (LLMs), built upon the Transformer architecture~\citep{vaswani2017attention}, have achieved remarkable progress in tasks such as machine translation, text classification, and generative dialogue. Despite these advances, their performance remains highly uneven across languages, favoring high-resource languages and marginalizing underrepresented ones~\citep{li2024quantifying,joshi-etal-2020-state,choudhury2023generative}. 
This imbalance deepens the digital language divide and limits the inclusivity of NLP technologies.

Recent work on Continual Pretraining (CPT) has shown promise for adapting pretrained models to new languages through additional training on targeted data~\citep{zheng2024breaking}.
EMMA-500 employed CPT with extensive monolingual datasets across more than 500 languages, significantly improving multilingual performance, particularly for low-resource languages~\citep{ji2024emma}. LLaMAX achieved notable translation improvements through CPT on over 100 languages involving data augmentation with bilingual translation data~\citep{lu-etal-2024-llamax}. 
Similar effects could be demonstrated on translation tasks with the CPT-based TOWER model~\citep{alves2024tower}.
However, the relative effectiveness of monolingual and bilingual translation data for CPT remains unclear, particularly in terms of their impact on continual language learning, language interference, and performance consistency across different resource levels of languages.

In addition to textual data in natural languages, a growing practice in LLM training is to incorporate programming code. For multilingual models, the inclusion of code, with its distinct logical syntax and lower ambiguity compared to natural language, can diversify the training data. We hypothesize that this diversity compels the model to learn more general and robust representations, moving beyond language-specific surface patterns; thus improving both the model's predictive ability and its performance on downstream tasks.

There is a critical gap in understanding how language characteristics interact with CPT strategies.
A recent classification proposed by \citet{yuan-etal-2024-vocabulary} categorizes languages as \textit{altruistic}, \textit{selfish}, and \textit{stagnant} based on their cross-lingual transfer patterns.
This framework classifies languages based on their impact in multilingual training: altruistic languages are hypothesized to improve performance on related languages, selfish languages primarily benefit themselves, and stagnant languages offer little to no benefit to any language, including their own.
However, this classification has only been validated in narrow experimental settings using English-centric bilingual data, leaving open questions about its generalizability to: (1) non-English language pairs, (2) code-augmented training regimes, and (3) models with varying pretraining corpora and architectures.

To systematically assess the impact of different CPT strategies,  we conduct extensive experiments with 36 configurations, evaluating monolingual, bilingual, and code-augmented CPT on multilingual adaptation.\footnote{Monolingual data consists of texts in a single language, though it may include code-switching. Bilingual translation data contains sentence pairs in two languages that convey the same meaning. When monolingual data from different languages is combined, it forms multilingual continual pertaining, and a similar principle applies to bilingual translation data. However, for clarity, we refer to these setups as monolingual CPT and bilingual CPT, respectively.}
Our setup includes three multilingual base models---Llama-3.1-8B \citep{dubey2024llama}, Llama-2-7B \citep{touvron2023llama}, and Viking-7B \citep{lumiopen_2025}---continual-pretrained languages spanning high-, medium-, and low-resource categories.
We evaluate the model performance on 14 training languages and assess cross-lingual transfer on 25 related languages, with a particular focus on assessing how different CPT configurations perform across altruistic, selfish, and stagnant language categories by \citet{yuan-etal-2024-vocabulary}. 

Our systematic evaluation of 36 CPT configurations across three base models and 30+ languages yields three core insights:  
\begin{itemize}
    \item \textbf{Bilingual CPT improves classification performance but introduces generation challenges:} Compared to monolingual CPT, bilingual CPT generally improves multilingual classification accuracy for medium- and low-resource languages. However, it frequently results in problematic language mixing during generation tasks, limiting its overall utility.
    
    \item \textbf{Code data enhances classification but introduces trade-offs in generation:}  Adding code data during CPT significantly boosts multilingual classification performance across resource levels, especially for low-resource languages, acting as an effective scaffold for representation learning. Nevertheless, code inclusion may lead to a trade-off, slightly degrading generation quality in certain scenarios.
    
    \item \textbf{The categorization of languages according to their cross-lingual transfer abilities does not generalize under varying conditions:}
    Our experiments reveal substantial deviations from language classifications proposed in previous work~\citep{yuan-etal-2024-vocabulary}:
    so-called {\em altruistic languages} are not always helpful and often negatively impact related languages, {\em selfish languages} exhibit highly configuration-dependent cross-lingual effects, and languages classified as {\em stagnant} demonstrate unexpected adaptability under specific training settings. These findings highlight the complexity of multilingual interactions in CPT and emphasize the need for a more adaptive classification framework for cross-lingual learning.
\end{itemize}

\section{Materials and Methods}
\subsection{Language Selection}
We systematically evaluate the effects of CPT on multilingual models by selecting languages according to the altruistic, selfish, and stagnant categories defined in \citet{yuan-etal-2024-vocabulary}, which classify languages based on their behavior in multilingual training and evaluation.

For each category, we select 1 high-resource language (except for the stagnant category, for which no high-resource language is available in the dataset we use), 2 medium-resource languages, and 2 low-resource languages to ensure a balanced representation across different resource levels. 
The classification of languages into high-, medium-, and low-resource categories is determined by analyzing the data distribution of the Lego-MT dataset~\citep{yuan-etal-2023-lego}, which serves as the basis for our setup. 
Specifically, we calculate the total token count for each language. 
Languages are then categorized as follows: high-resource languages exceed 1 billion tokens, medium-resource languages range between 10 million and 1 billion tokens, and low-resource languages fall below 10 million tokens.
These languages serve as the training languages in our CPT experiments. 
The selected training languages, along with their corresponding category and resource level, are summarized in the first three columns of Table~\ref{tab:combined-languages}.

To further validate the findings in \citet{yuan-etal-2024-vocabulary}, we select 1\text{-}2 linguistically related languages for each training language based on the language evolutionary tree\footnote{\url{http://www.elinguistics.net/Language_Evolutionary_Tree.html}}\footnote{Using the language evolutionary tree to identify related languages, we assess whether CPT effects transfer to unseen but linguistically similar languages, thus evaluating cross-lingual robustness.}. 
These related languages are not included in the CPT phase but are used for cross-lingual evaluation to determine whether the effects observed in training languages extend to unseen but related languages.  
The fourth and fifth columns of Table~\ref{tab:combined-languages} list the selected related languages. For some languages, this includes one, and for others, two related languages that are available in the evaluation benchmarks.

\begin{table}[htbp]
    \centering
    \small
    \begin{tabular}{llccc}
        \toprule
        \textbf{Category} & \textbf{Resources} & \textbf{Training Language} & \textbf{Related Language 1} & \textbf{Related Language 2} \\
        \midrule
        \multirow{5}{*}{Altruistic}  & High   & \texttt{zho\_Hani} & \texttt{yue\_Hant} & - \\
                                     & Medium & \texttt{ceb\_Latn} & \texttt{tgl\_Latn} & \texttt{ilo\_Latn} \\
                                     & Medium & \texttt{mar\_Deva} & \texttt{hin\_Deva} & \texttt{npi\_Deva} \\
                                     & Low    & \texttt{zul\_Latn} & \texttt{xho\_Latn} & \texttt{ssw\_Latn} \\
                                     & Low    & \texttt{khm\_Khmr} & \texttt{vie\_Latn} & - \\
        \hline
        \multirow{5}{*}{Selfish}     & High   & \texttt{deu\_Latn} & \texttt{nld\_Latn} & \texttt{dan\_Latn} \\
                                     & Medium & \texttt{bel\_Cyrl} & \texttt{rus\_Cyrl} & \texttt{ukr\_Cyrl} \\
                                     & Medium & \texttt{mri\_Latn} & \texttt{smo\_Latn} & \texttt{fij\_Latn} \\
                                     & Low    & \texttt{kir\_Cyrl} & \texttt{kaz\_Cyrl} & \texttt{bak\_Cyrl} \\
                                     & Low    & \texttt{nya\_Latn} & \texttt{bem\_Latn} & \texttt{sna\_Latn} \\
        \hline
        \multirow{4}{*}{Stagnant}    & Medium & \texttt{tha\_Thai} & \texttt{lao\_Laoo} & \texttt{shn\_Mymr} \\
                                     & Medium & \texttt{yor\_Latn} & \texttt{ibo\_Latn} & \texttt{hau\_Latn} \\
                                     & Low    & \texttt{sna\_Latn} & \texttt{nya\_Latn} & \texttt{zul\_Latn} \\
                                     & Low    & \texttt{wol\_Latn} & \texttt{bam\_Latn} & - \\
        \bottomrule
    \end{tabular}
    \caption{Selected languages for CPT along with their corresponding related languages for evaluation. `-' indicates the second related language cannot be found in the benchmark.}
    \label{tab:combined-languages}
\end{table}

\subsection{Pretraining Data}
\label{subsec:pt_data}
\paragraph{Bilingual Translation Data}
We utilize subsets of the Lego-MT~\citep{yuan-etal-2023-lego} and NLLB~\citep{schwenk-etal-2021-ccmatrix,heffernan2022bitext,costa2022no} datasets as our sources of parallel bilingual data.  
The Lego-MT dataset, derived from OPUS\footnote{\url{https://opus.nlpl.eu}}, provides translations across 433 languages.  
The NLLB dataset consists of 148 English-centric and 1,465 non-English-centric bitext pairs mined from different parallel sources.
To construct our parallel training data, we select specific language pairs from these datasets and apply OpusFilter~\citep{aulamo-etal-2020-opusfilter} to remove duplicate data points.

The resulting dataset comprises approximately 292 million tokens across 22 language pairs, distributed over three language categories: altruistic (10 pairs, \textasciitilde 92M tokens), selfish (8 pairs, \textasciitilde 100M tokens), and stagnant (4 pairs, \textasciitilde 100M tokens).

For training, we format parallel data using the following structure:

\noindent \hfill
\texttt{\textcolor{blue}{[source language]}: [source] \textcolor{blue}{[target language]}: [target]}
\hfill{}

\paragraph{Monolingual Data}
We extract a subset of MADLAD-400~\citep{kudugunta2024madlad}, a large-scale multilingual dataset derived from Common Crawl\footnote{\url{https://commoncrawl.org/}}, covering 419 languages.
Since web-crawled text does not inherently guarantee monolingual integrity, we employ GlotLID~\citep{kargaran-etal-2023-glotlid}, a language identification model, to analyze the language composition of each text segment and ensure strict monolingual consistency. 
Specifically, for each document in the dataset, we first segment the text into sentences using the NLTK~\citep{bird-loper-2004-nltk} sentence splitter. 
Then, GlotLID predicts the language of each sentence independently. 
We retain only those documents where all sentences are identified as belonging to the same language, discarding any text segment that exhibits code-switching or multilingual content.

We finally select data for 15 unique languages, with English appearing in all three categories: altruistic (6 languages, \textasciitilde 92M tokens), selfish (6 languages, \textasciitilde 100M tokens), and stagnant (5 languages, \textasciitilde 87M tokens), resulting in a total of approximately 279 million tokens.

\paragraph{Code Data}
We incorporate code data from The Stack~\citep{kocetkov2022stack}, following the pre-processing strategy used in EMMA-500~\citep{ji2024emma}. 
The dataset is first filtered to retain high-quality source files, with a focus on data science-related code and the 32 most commonly used general-purpose programming languages. 
Additionally, we include LLVM code due to its importance in multilingual code generation~\citep{paul-etal-2024-ircoder,szafraniec2022code}.

For training configurations that include code, we establish a 2:1 ratio between textual (100 million tokens) and code data (50 million tokens), with code comprising about 33\% of the total mixture. 
We intentionally selected this higher proportion to investigate the effects of a strong code signal within our CPT context.
While prior work~\citep{aryabumi2024code} on pretraining from scratch recommends a 25\% proportion for balancing language and code performance, it is noted that a higher 33\% remains reasonable for enhancing models’ ability, aligning with our goal of studying its trade-offs.

\subsection{Base Models}
We evaluate across three open-source multilingual LLMs with diverse training recipes:

Llama-3.1-8B~\citep{dubey2024llama} is pretrained on approximately 15 trillion tokens from diverse, multilingual sources. Its extensive multilingual pretraining and high capacity make it ideal for analyzing CPT effects on well-trained models.

Llama-2-7B~\citep{touvron2023llama} is pretrained on 2 trillion tokens, covering a broad yet less multilingual data distribution. It provides a baseline to evaluate CPT effectiveness on English-centric models commonly used in multilingual adaptation research.

Viking-7B~\citep{lumiopen_2025} is pretrained mainly on Nordic languages, English, and code, offering insights into how CPT impacts models initially trained on narrower, region-specific data.

\subsection{CPT Configurations}
We train models under 4 CPT configurations across 3 base models and 3 language categories, resulting in a total of 36 models. Each model is named using the format:
\[
\texttt{Model-Data[+Code]-LangCat}
\]

where:
\begin{itemize}
    \item \texttt{Model} $\in \{ \text{L3 (Llama-3.1-8B)}, \text{L2 (Llama-2-7B)}, \text{V7 (Viking-7B)} \}$
    \item \texttt{Data} $\in \{ \text{Mono (Monolingual)}, \text{Bi (Bilingual)} \}$
    \item \texttt{Code} (optional) is added if code data is included
    \item \texttt{LangCat} $\in \{ \text{Alt (Altruistic)}, \text{Sel (Selfish)}, \text{Stag (Stagnant)} \}$
\end{itemize}

For example, \texttt{L3-Mono-Alt} refers to Llama-3.1-8B trained on monolingual data for altruistic languages, while \texttt{L2-Bi+Code-Sel} denotes Llama-2-7B trained on bilingual parallel texts in selfish languages and code data.

Each model is trained for 2 epochs on a cluster with 4 $\times$ AMD MI250X GPUs (8 Graphics Compute Dies) on each node. 
Training data is organized by language category (altruistic, selfish, stagnant), with all languages within a category (e.g., altruistic: \texttt{zho\_Hani}, \texttt{ceb\_Latn}, etc.) mixed into a single dataset per configuration (e.g., monolingual, bilingual+code).
As for software, we use the LLaMA-Factory~\citep{zheng2024llamafactory} framework with DeepSpeed~\citep{rajbhandari2020zero} ZeRO-3 config. The hyperparameter setup includes a per-device batch size of 8 with gradient accumulation steps of 2. We use a cosine learning rate scheduler with an initial learning rate of $4.0 \times 10^{-5}$ and a warmup ratio of 0.03.

\section{Evaluation and Discussion}
\subsection{Benchmarks and Setup}
We evaluate our models on two highly multilingual benchmarks covering a classification and a generation task:
SIB-200~\citep{adelani-etal-2024-sib} for topic classification and FLORES-200~\citep{costa2022no,goyal-etal-2022-flores,guzman-etal-2019-flores} for machine translation. 
Classification focuses on whether CPT improves the multilingual model's understanding within a single language, while translation studies the alignment between languages that emerges with multilingual CPT.
All experiments use a consistent 3-shot prompting setup.

\paragraph{SIB-200}
SIB-200 is a multilingual news topic classification benchmark covering 200 languages. The task involves classifying news headlines into one of the following predefined categories: science/technology, travel, politics, sports, health, entertainment, and geography.

The model predicts by ranking logits for each category, and accuracy measures performance across languages.

\paragraph{FLORES-200}
FLORES-200 evaluates multilingual translation performance across diverse language pairs.

Translations are generated using the vLLM~\citep{kwon2023efficient} inference engine. BLEU~\citep{papineni2002bleu} scores, computed via \texttt{SacreBLEU}~\citep{post2018call} with the flores200 tokenizer, quantify translation quality. \footnote{BLEU signature: \texttt{nrefs:1|case:mixed|eff:no|tok:flores200|smooth:exp|version:2.4.2}}

\subsection{Effect of Monolingual and Bilingual Continual Pretraining}

This section shows that bilingual CPT hampers generation due to language mixing but excels in classification for medium- and low-resource languages over monolingual CPT.

\subsubsection{Language Mixing in Generation Tasks}
\label{sec:language_mixing}
The FLORES-200 translation task revealed significant language mixing issues in models trained with bilingual translation data. 
Specifically, when generating translations between language pairs, models frequently appended unintended language tokens to the output. 
For example, when translating from English (eng\_Latn) to Chinese (\texttt{zho\_Hani}), models trained on bilingual data produced outputs like:
\begin{quote}
\begin{center}
    \vspace{-10pt}
    \includegraphics[width=0.8\textwidth]{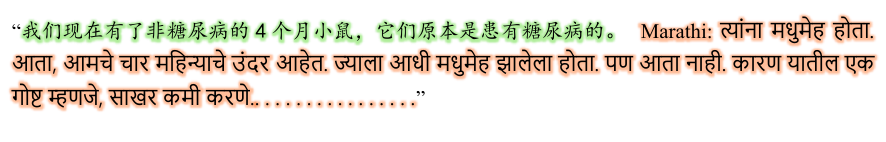}
    \vspace{-10pt}
\end{center}
\end{quote}

The text with a green background represents the desired Chinese translation, while the text with an orange background contains nonsensical multilingual fragments. This phenomenon occurred consistently across bilingual CPT configurations, suggesting that the parallel data format (\texttt{[Lang1]: xxx [Lang2]: yyy}) encourages cross-lingual interference. More examples are in \Cref{fig:language_mixing_examples} in \Cref{sec:more_language_mixing}.

\begin{figure}[htbp]
    \centering
    \includegraphics[width=\textwidth]{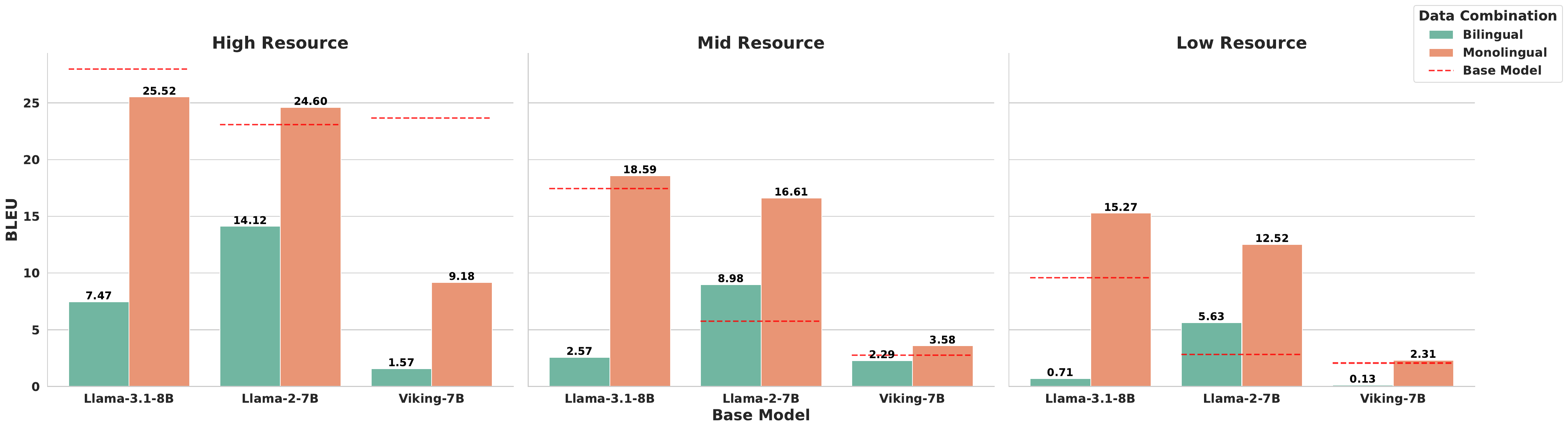}
    \caption{FLORES-200 X-Eng BLEU score comparing bilingual and monolingual CPT across high-, mid-, and low-resource languages.}
    \label{fig:flores200_bi_vs_mo}
\end{figure}

This language inconsistency leads to significant translation quality degradation, as shown in Figure~\ref{fig:flores200_bi_vs_mo}. Bilingual CPT configurations underperform monolingual CPT across all resource levels and base models. For high-resource languages, Llama-3.1-8B achieves only 7.47 BLEU with bilingual CPT versus 25.52 with monolingual CPT (-71\% relative), while Llama-2-7B shows similar disparities (14.12 vs 24.60, -43\%). The pattern persists for mid- and low-resource languages, with bilingual CPT consistently lagging behind monolingual CPT. Notably, monolingual CPT often matches or exceeds baseline performance, whereas bilingual CPT only exceeds baseline in specific cases, such as Llama-2-7B on mid- and low-resource languages.
\Cref{sec:detailed_results_flores200} presents the detailed results on each language.

\subsubsection{Comparative Analysis in Classification Tasks}
\begin{figure}[htbp]
    \centering
    \includegraphics[width=\textwidth]{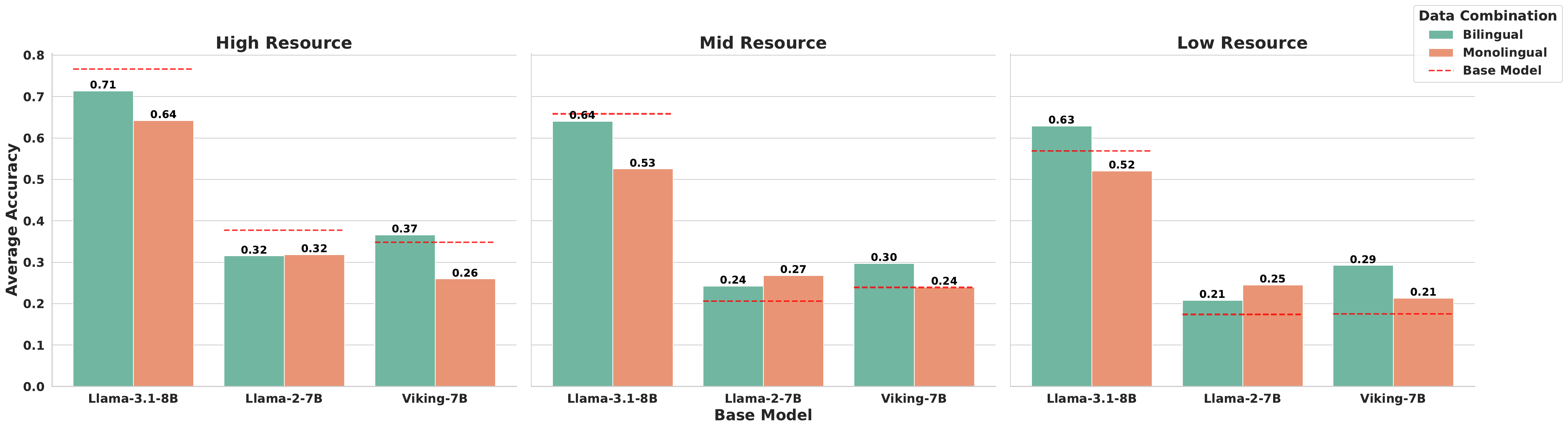}
    \caption{SIB-200 classification accuracy comparing monolingual and bilingual CPT across high-, mid-, and low-resource languages.}
    \label{fig:sib200_bi_vs_mo}
\end{figure}

To isolate the effects of CPT strategies without interference from language mixing, we evaluate SIB-200 classification accuracy. 
Figure \ref{fig:sib200_bi_vs_mo} shows the average accuracy aggregated across models trained separately on altruistic, selfish, and stagnant languages, grouped by resource level.

\paragraph{High-Resource Languages}
For high-resource languages, both monolingual and bilingual CPT degrade performance across all base models compared to their respective baselines. Llama-3.1-8B, despite its strong baseline (76.63\%), exhibits drops with bilingual CPT (71.41\%, -6.8\% relative) and monolingual CPT (64.21\%, -16.2\%).
Llama-2-7B shows significant declines with both strategies: bilingual CPT reduces accuracy to 31.54\% (vs baseline 37.75\%, -16.5\%), while monolingual CPT performs similarly (31.86\%, -15.6\%).
Viking-7B partially escapes this trend, with bilingual CPT achieving marginal gains (36.60\% vs baseline 34.80\%, +5.2\%), though monolingual CPT underperforms (25.98\%, -25.3\%). 
This suggests that high-resource languages generally do not benefit from CPT, likely due to interference with existing strong representations in pretrained models. 
However, model-specific factors, such as whether the model's pretraining data aligns well with the target languages in CPT, may enable limited improvements in certain cases. 
For example, Viking-7B, which was pretrained primarily on Nordic languages and English, may benefit more from bilingual CPT due to its ability to leverage cross-lingual transfer between related languages.

\paragraph{Mid-Resource Languages}
Mid-resource languages show mixed trends. 
Llama-3.1-8B maintains near-baseline performance with bilingual CPT (64.05\% vs baseline 65.85\%, -2.7\%), but monolingual CPT degrades significantly (52.53\%, -20.2\%). 
Llama-2-7B improves accuracy in both configurations: bilingual CPT raises accuracy to 24.26\% (vs. baseline 20.59\%, +17.8\%), while monolingual CPT performs slightly better at 26.80\% (+30.2\%).
Viking-7B uniquely benefits from bilingual CPT (29.74\% vs baseline 23.94\%, +24.2\%), while monolingual CPT underperforms (24.02\%, +0.3\%). 
This indicates that bilingual CPT can stabilize mid-resource language performance for certain models (e.g., Viking-7B and Llama-2-7B). 
However, monolingual CPT risks overfitting to limited in-language data, particularly for models with weaker pretraining (e.g., Llama-3.1-8B).

\paragraph{Low-Resource Languages}
Low-resource languages exhibit divergent patterns. 
Llama-3.1-8B improves with bilingual CPT (62.91\% vs baseline 56.86\%, +10.6\%) but declines with monolingual CPT (52.04\%, -8.5\%). 
Llama-2-7B improves moderately with bilingual CPT (20.84\%, +19.8\%) and shows stronger gains with monolingual CPT (24.51\%, +40.9\%). 
Viking-7B benefits substantially from bilingual CPT (29.25\%, +66.5\%), while monolingual CPT slightly underperforms (21.33\%, +21.4\%).
This highlights that bilingual CPT can enhance low-resource language performance for models with compatible pretraining (e.g., Viking-7B and Llama-3.1-8B).

\subsection{Effect of Including Code Data}
\label{sec:effect_code}
\subsubsection{Downstream Tasks Performance}
\begin{figure}[htbp]
    \centering
    \includegraphics[width=\textwidth]{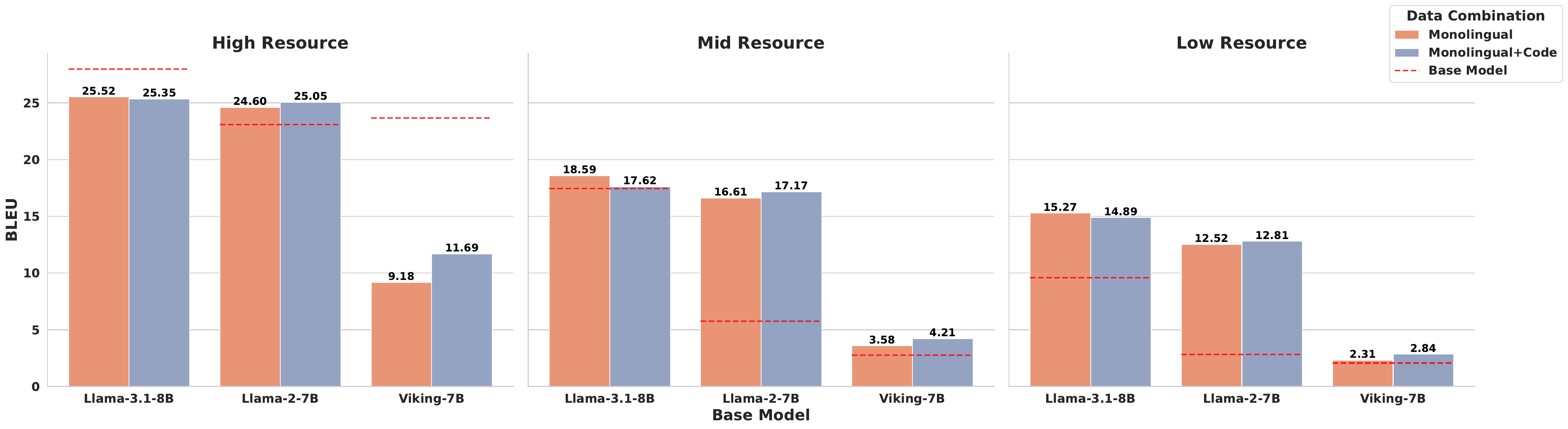}
    \caption{FLORES-200 X-Eng BLEU score comparing monolingual and monolingual+code CPT across high-, mid-, and low-resource languages.}
    \label{fig:flores200_mo_vs_mo+code}
\end{figure}

\begin{figure}[htbp]
    \centering
    \includegraphics[width=\textwidth]{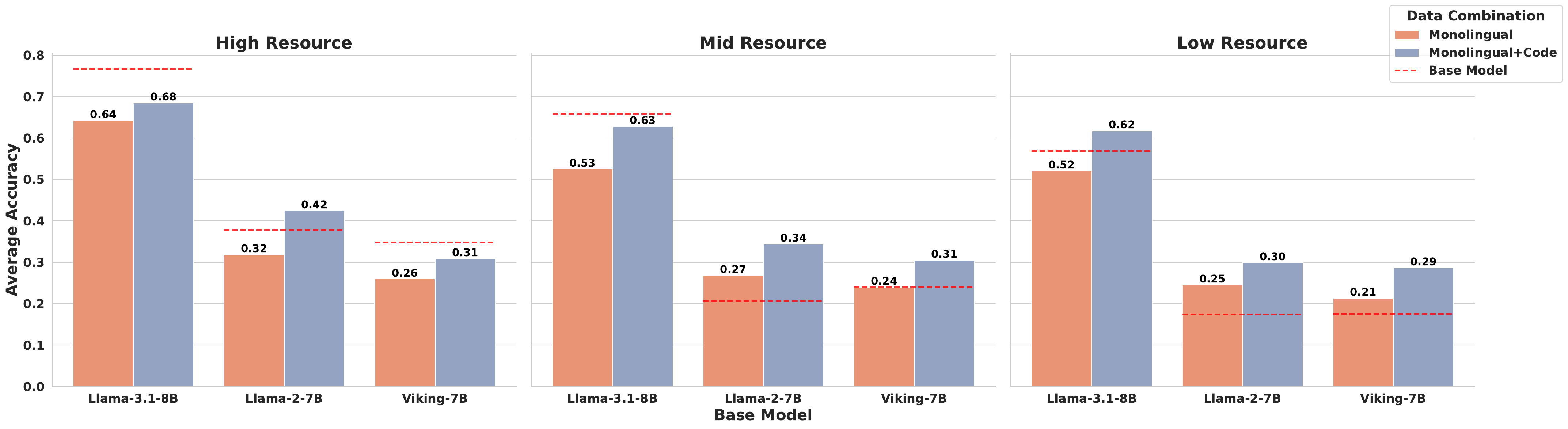}
    \caption{SIB-200 classification accuracy comparing monolingual and monolingual+code CPT across high-, mid-, and low-resource languages.}
    \label{fig:sib200_mo_vs_mo+code}
\end{figure}

The integration of code data during monolingual CPT shows task-dependent effects, enhancing classification performance while introducing tradeoffs in generation quality. Figure \ref{fig:sib200_mo_vs_mo+code} and Figure \ref{fig:flores200_mo_vs_mo+code} compare monolingual CPT with and without code data across resource levels and tasks, revealing key patterns in how code data influences multilingual adaptation.

Code integration consistently improves classification accuracy across all resource levels and models. 
For high-resource languages, Llama-3.1-8B shows marginal gains (64.21\% to 68.47\%, +6.7\% relative to baseline 76.63\%), while Llama-2-7B and Viking-7B exhibit more substantial improvements (42.48\% vs 31.86\%, +33.3\%; 30.88\% vs 25.98\%, +18.8\%).
Mid-resource languages benefit even more, with Llama-3.1-8B recovering near-baseline performance (52.53\% to 62.83\%, -4.6\% vs baseline 65.85\%) and Llama-2-7B achieving significant gains (34.40\% vs 26.80\%, +67.0\%).
Low-resource languages see the most pronounced improvements, particularly for Viking-7B (28.68\% vs 21.33\%, +63.2\% relative to baseline 17.57\%). 
This pattern extends to bilingual CPT configurations (see \Cref{sec:bilingual_code_appendix}).

In contrast, code integration often degrades translation quality, particularly for high-resource languages. 
Llama-3.1-8B shows slight degradation (25.52 BLEU to 25.35, -0.7\% vs baseline 27.97), while Llama-2-7B and Viking-7B exhibit gains (25.05 vs 24.60, +8.6\%; 11.69 vs 9.18, +27.3\%). 
Mid-resource languages show mixed trends, with Llama-3.1-8B experiencing a slight drop (17.62 vs 18.59, -5.2\%) and Viking-7B improving significantly (4.21 vs 3.58, +52.0\%). 
Low-resource languages partially escape this trend, with Viking-7B showing substantial gains (2.84 vs 2.31, +37.4\%).

The benefits of code integration are most pronounced for low-resource languages, where it acts as a "scaffold" to improve classification accuracy (avg. +25.1\%) and partially mitigate generation deficits. Mid-resource languages also benefit, though to a lesser extent, while high-resource languages see diminishing returns, with classification gains (e.g., Llama-3.1-8B: +6.7\%) offset by generation losses.

\subsubsection{Language Modeling Performance}
To assess the impact of code integration on language modeling, we evaluate performance using Negative Log-Likelihood (NLL) on the MaLA validation set\footnote{\url{https://huggingface.co/datasets/MaLA-LM/mala-monolingual-split}}. 
We use the implementation of GlotEval \citep{gloteval}, where the NLL is calculated by concatenating input sentences and using a strided sliding window of 1024 tokens, and is defined as $$NLL=-\sum_{i=1}^{n_i}log p_{\theta}(x_i|x_{<i}).$$
Our results are summarized in Table~\ref{tab:nll_mo_vs_mo+code}. Across all three base models and all resource levels, the Monolingual+Code configurations yielded lower (better) NLL scores than the monolingual-only CPT. This demonstrates that including code not only helps in downstream classification tasks but also improves the models' predictive ability and likelihood estimation of unseen text during CPT.

\begin{table}[htbp]
\centering
\tiny
\renewcommand{\arraystretch}{0.8}
\resizebox{0.75\textwidth}{!}{
\begin{tabular}{lccc}
\toprule
\textbf{Model} & \textbf{High Resource} & \textbf{Mid Resource} & \textbf{Low Resource} \\
\midrule
L2-Mono-       & 5.0080 & 1.7609 & 4.3308 \\
L2-Mono+Code-  & \textbf{4.7359} & \textbf{1.7042} & \textbf{4.2289} \\
\midrule
L3-Mono-       & 4.8864 & 2.0375 & 4.2013 \\
L3-Mono+Code-  & \textbf{4.5701} & \textbf{1.9522} & \textbf{4.1389} \\
\midrule
V7-Mono-       & 6.9916 & 1.9506 & 4.5367 \\
V7-Mono+Code-  & \textbf{6.6029} & \textbf{1.8369} & \textbf{4.4850} \\
\bottomrule
\end{tabular}
}
\caption{NLL across different models and resource levels. Lower is better.}
\label{tab:nll_mo_vs_mo+code}
\end{table}

\subsection{Validation of Language Category Hypotheses}
This section evaluates the validity of the altruistic, selfish, and stagnant language classifications proposed in prior work~\citep{yuan-etal-2024-vocabulary}.
We evaluate each model (e.g., \texttt{L3-Mono-Alt}) trained on a language category (e.g., altruistic languages: \texttt{zho\_Hani}, \texttt{ceb\_Latn}, etc.) and measure SIB-200 classification accuracy changes on both the trained languages and their related languages (e.g., \texttt{yue\_Hant}, \texttt{tgl\_Latn}, etc.), as defined in Table~\ref{tab:combined-languages}.
We analyze whether CPT strategies align with these hypothesized behaviors. Table~\ref{tab:val-related} reports accuracy changes (\%) relative to base models.

\begin{table}[htbp]
\centering
\scriptsize 
\resizebox{\textwidth}{!}{
\begin{tabular}{l|lll|lll|lll}
\toprule
        \multirow{2}{*}{\textbf{Model}} & \multicolumn{3}{c|}{\textbf{Altruistic Languages}} 
                               & \multicolumn{3}{c|}{\textbf{Selfish Languages}} 
                               & \multicolumn{3}{c}{\textbf{Stagnant Languages}} \\
                       & Training      & Related      & Met?      & Training     & Related     & Met?     & Training      & Related     & Met?     \\
\midrule
L2-Bi-         & +7.08  & -22.55  & \textcolor{red}{No}  & +12.33  & +2.90   & \textcolor{green}{Yes}  & +5.88   & -9.99   & \textcolor{red}{No} \\
L2-Bi+Code-    & +62.37 & +28.31  & \textcolor{red}{No}  & +52.32  & +31.67  & \textcolor{red}{No}  & +26.13  & +6.25   & \textcolor{red}{No} \\
L2-Mono-       & -14.60 & -31.32  & \textcolor{red}{No}  & +53.18  & +21.94  & \textcolor{red}{No}  & +31.36  & -8.33   & \textcolor{red}{No} \\
L2-Mono+Code-  & +50.43 & +19.04  & \textcolor{red}{No}  & +52.32  & +26.29  & \textcolor{red}{No}  & +64.02  & +14.57  & \textcolor{red}{No} \\
\hline
L3-Bi-         & +4.46  & -4.46   & \textcolor{red}{No}  & -7.90   & -19.54  & \textcolor{red}{No}  & +14.76  & -28.43  & \textcolor{red}{No} \\
L3-Bi+Code-    & +1.64  & -7.70   & \textcolor{red}{No}  & -5.85   & -15.66  & \textcolor{red}{No}  & +21.81  & -28.04  & \textcolor{red}{No} \\
L3-Mono-       & -24.37 & -31.26  & \textcolor{red}{No}  & -9.07   & -19.84  & \textcolor{red}{No}  & -7.71   & -43.54  & \textcolor{red}{No} \\
L3-Mono+Code-  & -1.78  & -11.13  & \textcolor{red}{No}  & +2.49   & -10.85  & \textcolor{red}{No}  & 0.00    & -37.01  & \textcolor{red}{No} \\
\hline
V7-Bi-         & -11.41 & -31.95  & \textcolor{red}{No}  & +19.24  & -10.22  & \textcolor{red}{No}  & +78.18  & +26.32  & \textcolor{red}{No} \\
V7-Bi+Code-    & +22.82 & -9.35   & \textcolor{red}{No}  & +17.57  & -16.35  & \textcolor{red}{No}  & +11.16  & -19.36  & \textcolor{red}{No} \\
V7-Mono-       & -8.22  & -19.74  & \textcolor{red}{No}  & +5.86   & -33.45  & \textcolor{red}{No}  & 0.00    & -0.83   & \textcolor{green}{Yes} \\
V7-Mono+Code-  & +5.93  & -11.69  & \textcolor{red}{No}  & +53.96  & +8.18   & \textcolor{green}{Yes}  & +21.31  & +17.27  & \textcolor{red}{No} \\
\bottomrule
\end{tabular}
}
\caption{SIB-200 classification accuracy changes (\%) for training and related languages across altruistic, selfish, and stagnant categories. Results are reported relative to base models, with a ``Met'' column to indicate whether the hypothesis is met or contradicted.}
\label{tab:val-related}
\end{table}

\paragraph{Altruistic languages can also be selfish or mutually harmful} %
The altruistic hypothesis indicates that training in altruistic languages enhances multilingual performance (related languages) with minimal impact on their own performance (trained languages).
Our results reveal three critical contradictions: 
(1) 83\% of configurations (10/12) degraded related language performance, with code-free CPT causing up to -31.32\% accuracy (L2-Mono-Alt); 
(2) Contrary to "minimal self-impact", trained language accuracy fluctuated wildly (+62.37\% in L2-Bi+Code-Alt vs. -24.37\% in L3-Mono-Alt); 
These bidirectional effects challenge the unidirectional altruism assumption.

\paragraph{Selfish languages exhibit conditional isolation only in certain cases}
While the selfish hypothesis suggests trained languages primarily improve their own performance (trained languages) while minimally affecting others (related languages), we find this only holds in specific configurations: 
(1) Non-code bilingual training (L2-Bi-Sel) showed minimal impact on related languages (+2.90\%); 
(2) Code-augmented monolingual training (V7-Mono+Code-Sel) achieved strong self-improvement (+53.96\%) with moderate spillover (+8.18\%). 
However, 83\% of cases (10/12) violated the hypothesis through either negative spillover (V7-Mono-Sel: -33.45\%) or excessive cross-lingual transfer (L2-Bi+Code-Sel: +31.67\%). 

\paragraph{Stagnant languages demonstrate more adaptability than expected}
Stagnant languages neither improve their own performance (trained languages) nor influence others 
Contrary to their purported stagnation, 92\% of configurations (11/12) induced significant performance shifts: 
(1) Bilingual training boosted trained languages by +78.18\% (V7-Bi-Stag) while improving related languages (+26.13\%); 
(2) Monolingual+code CPT (L2-Mono+Code-Stag) achieved +64\% self-improvement with +14.76\% cross-lingual gains. 
Only V7-Mono-Stag showed true stagnation (+0.00\% trained, -0.83\% related). 
This reveals that most ``stagnant'' languages possess untapped adaptation potential under proper CPT strategies.

\section{Conclusion}
In this study, we systematically evaluated the effects of multilingual CPT strategies, including monolingual, bilingual, and code-augmented configurations, across diverse resource levels and language categories. Through experiments with 36 configurations involving three multilingual base models and over 30 languages, we identified several critical insights:

First, while bilingual CPT enhances classification accuracy for mid- and low-resource languages, it introduces language mixing during generation, limiting its utility for translation tasks. 
Second, code integration during CPT acts as a scaffold for low-resource language understanding but introduces task-dependent trade-offs, improving classification and language modeling capabilities while slightly degrading generation quality. 
Third, we demonstrate that language classifications based on cross-lingual transfer patterns (\textit{altruistic, selfish, stagnant}) fail to generalize under varying CPT strategies.

Overall, our work underscores the complexity of multilingual representation learning and highlights the need for flexible frameworks for language categorization and training strategy selection. Future research should focus on developing more adaptive CPT methods that balance classification improvements and generation quality, further bridging language disparities in large language models.

\section*{Ethics Statement}
This research focuses on reducing the digital language divide and improving inclusivity for underrepresented languages. We acknowledge potential biases due to uneven data distribution and strive to mitigate them by including diverse languages across resource levels. All datasets used are publicly available and preprocessed to ensure integrity and monolingual consistency. 
All the models trained in this paper are strictly for research purposes and are not intended to be deployed in real-world applications. 
We encourage further work to address ethical challenges in multilingual NLP, especially for underrepresented languages.

\section*{Reproducibility Statement}
To ensure reproducibility, we release:
\begin{itemize}
\item \textbf{Model Checkpoints:} All models trained under various configurations (monolingual, bilingual, code-augmented) across base models (Llama-3.1-8B, Llama-2-7B, Viking-7B) and language categories (altruistic, selfish, stagnant).
\item \textbf{Processed Dataset:} Filtered subsets of Lego-MT, NLLB, MADLAD-400, and code data.
\item \textbf{Scripts:} Data cleaning, training, and evaluation scripts, including LLaMA-Factory with DeepSpeed ZeRO-3 configuration.
\end{itemize}
All resources are available at: \url{https://mala-lm.github.io/MixCPT.html}.

\section*{Acknowledgments}
The work has received funding from the Digital Europe Programme under grant agreement No 101195233 (OpenEuroLLM). The authors wish to acknowledge CSC - IT Center for Science, Finland, for providing computational resources.

\bibliography{colm2025_conference}
\bibliographystyle{colm2025_conference}

\appendix
\section{Appendix}
\subsection{Related Work}
\paragraph{Continual Pretraining and Language Adaptation}
Continual pretraining has emerged as a pivotal technique for adapting LLMs to new domains or languages while retaining previously acquired knowledge \citep{yildiz2024investigating}. This approach has demonstrated significant benefits across diverse domains, including cybersecurity \citep{yu2025primus}, finance\citep{hirano2024construction}, and law \citep{niyogi2024paramanu}.
In the context of language adaptation, researchers have successfully leveraged continual pretraining to enhance performance on low- and medium-resource languages. 
For instance, \citet{ji2024emma, ji2025massively, lu-etal-2024-llamax} extended the capabilities of open-weight LLMs by pretraining them on multilingual datasets encompassing hundreds of languages. 
Similarly, \citet{fujii2024continual} significantly improved Japanese language proficiency by continually pretraining LLama-2~\citep{touvron2023llama} on a large-scale Japanese web corpus. \citet{vo2024redwhale} achieved notable advancements in Korean language processing by utilizing 9.7 billion tokens for continual pretraining.
Beyond direct continual pretraining, \citet{yong-etal-2023-bloom} benchmarks various language adaptation strategies, including adapter-based tuning (MAD-X~\citep{pfeiffer2020mad}), and parameter-efficient methods like (IA)³~\citep{liu2022few}, applying them to the BLOOM model under low-resource constraints.
Babel~\citep{zhao2025babel} introduces a structural expansion of LLMs via layer extension and trains them on curated multilingual corpora, thereby enabling support for 25 widely spoken but underrepresented languages.

\paragraph{Bilingual Translation Data}
Incorporating bilingual translation data into pretraining has been shown to enhance multilingual performance, although the benefits tend to diminish as model size increases~\citep{kale-etal-2021-nmt5}. 
Even relatively small parallel corpora, such as 10,000 sentence pairs, can be as effective as much larger datasets when carefully filtered for quality~\citep{lin2024recipe}.
Recent efforts further highlight how strategically leveraging bilingual data can enhance multilingual capabilities. 
For example, \citet{ranaldi-etal-2024-empowering} introduced \textit{Translation-following} demonstrations to improve semantic alignment between English and other languages during instruction tuning. Their CrossAlpaca models, trained with both instruction and translation data, significantly outperformed monolingual baselines on multilingual QA tasks.
Similarly, \citet{alves2024tower} showed that including high-quality parallel data during continual pretraining, alongside monolingual data, leads to substantial improvements in translation and related tasks.
In contrast to the improvement from training with bilingual translation data, \citet{ji2024can} found that utilizing bilingual translation to enforce sentence-level alignment during continual pretraining actually hinders cross-lingual transfer based on the study on mBART \citep{tang-etal-2021-multilingual}.

\paragraph{Code Data in Language Model Training}
Including code in pretraining data has become a common practice, even for models not specifically designed for code generation~\citep{chen2021evaluating}. 
Recent studies show that code data not only improves performance on programming tasks but also enhances general capabilities such as natural language reasoning, entity tracking, and commonsense understanding~\citep{aryabumi2024code}. 
For instance, models trained with code exhibit stronger performance in structured reasoning tasks~\citep{madaan2022language} and demonstrate better entity tracking compared to purely text-trained counterparts~\citep{kim2024code}. 
Furthermore, adding high-quality or synthetic code during pretraining or cooldown leads to consistent gains across a wide range of benchmarks~\citep{aryabumi2024code}. 
Systematic experiments also suggest that mixing code data during both pretraining and instruction tuning stages leads to better reasoning abilities without harming performance on non-code tasks~\citep{ma2023training}.

\paragraph{Challenges and Dynamics of Cross-Lingual Transfer}
While multilingual training can confer benefits, it also introduces the risk of performance degradation, a phenomenon termed the "curse of multilinguality" or "negative interference". \citet{chang-etal-2024-multilinguality} systematically investigated multilingual pretraining and revealed that adding multilingual data can improve performance for low-resource languages up to a certain threshold, beyond which the performance degrades for both low-resource and high-resource languages due to limited model capacity. Moreover, \citet{wang-etal-2020-negative} further demonstrated that negative interference impacts not only high-resource but also low-resource languages, highlighting the universality of this challenge in multilingual modeling.
However, multilingual training also presents significant opportunities, particularly through cross-lingual transfer. \citet{protasov-etal-2024-super} demonstrates that certain high-resource languages, termed "super donors," can effectively boost the performance of low-resource "super recipient" languages in multilingual models.

\subsection{CPT Configurations with Structured Naming}
\begin{table}[htbp]
\centering
\scriptsize
\setlength{\tabcolsep}{3pt}
\renewcommand{\arraystretch}{1.2}
\begin{tabular}{llcccc}
\toprule
\multirow{2}{*}{\textbf{Base Model}} & \multirow{2}{*}{\textbf{Category}} & \multicolumn{4}{c}{\textbf{Training Data}} \\
\cline{3-6}
 &  & Mono & Bi & Mono+Code & Bi+Code \\
\midrule

\multirow{3}{*}{Llama-3.1-8B} 
 & Altruistic  & L3-Mono-Alt  & L3-Bi-Alt  & L3-Mono+Code-Alt  & L3-Bi+Code-Alt  \\
 & Selfish     & L3-Mono-Sel  & L3-Bi-Sel  & L3-Mono+Code-Sel  & L3-Bi+Code-Sel  \\
 & Stagnant    & L3-Mono-Stag & L3-Bi-Stag & L3-Mono+Code-Stag & L3-Bi+Code-Stag \\
\hline

\multirow{3}{*}{Llama-2-7B}
 & Altruistic  & L2-Mono-Alt  & L2-Bi-Alt  & L2-Mono+Code-Alt  & L2-Bi+Code-Alt  \\
 & Selfish     & L2-Mono-Sel  & L2-Bi-Sel  & L2-Mono+Code-Sel  & L2-Bi+Code-Sel  \\
 & Stagnant    & L2-Mono-Stag & L2-Bi-Stag & L2-Mono+Code-Stag & L2-Bi+Code-Stag \\
\hline

\multirow{3}{*}{Viking-7B}
 & Altruistic  & V7-Mono-Alt  & V7-Bi-Alt  & V7-Mono+Code-Alt  & V7-Bi+Code-Alt  \\
 & Selfish     & V7-Mono-Sel  & V7-Bi-Sel  & V7-Mono+Code-Sel  & V7-Bi+Code-Sel  \\
 & Stagnant    & V7-Mono-Stag & V7-Bi-Stag & V7-Mono+Code-Stag & V7-Bi+Code-Stag \\
\bottomrule

\end{tabular}
\label{tab:cpt-models}
\end{table}

\subsection{Additional Results on Bilingual CPT with Code Data}
\label{sec:bilingual_code_appendix}
\begin{figure}[htbp]
    \centering
    \includegraphics[width=\textwidth]{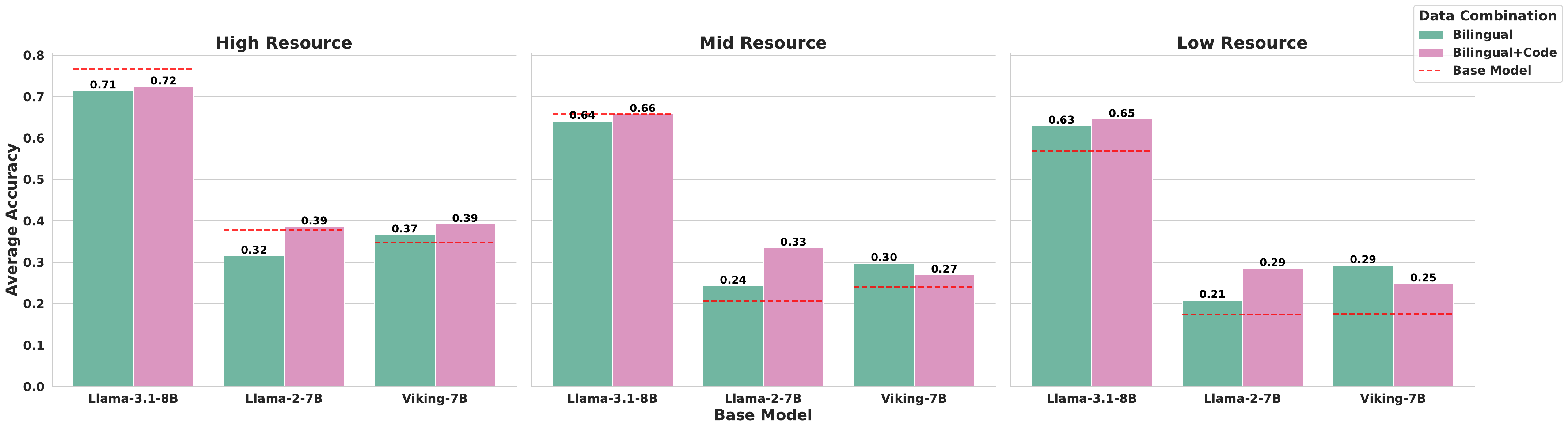}
    \caption{SIB-200 classification accuracy comparing bilingual and bilingual+code CPT across high-, mid-, and low-resource languages.}
    \label{fig:sib200_bi_vs_bi+code}
\end{figure}

Figure \ref{fig:sib200_bi_vs_bi+code} shows the impact of adding code data to bilingual CPT configurations for the SIB-200 classification task. 
While \Cref{sec:effect_code} in the main text focuses on monolingual CPT comparisons, the results in this section demonstrate that code integration also benefits bilingual CPT across most models and language resource levels for natural language understanding. 
For high-resource languages, the improvements are modest but consistent: Llama-3.1-8B increases from 71.41\% to 72.39\% (+1.4\% relative), Viking-7B from 36.60\% to 39.21\% (+7.1\%), and Llama-2-7B shows the largest gain (31.54\% to 38.56\%, +22.3\%). 
Mid-resource languages exhibit similar patterns, with Llama-2-7B improving from 24.26\% to 33.50\% (+38.0\%) and Llama-3.1-8B from 64.05\% to 65.77\% (+2.7\%). Notably, Viking-7B shows a slight degradation (29.74\% to 26.96\%, -9.3\%), suggesting model-specific sensitivities to code interference in this configuration.
The most significant benefits emerge for low-resource languages, Llama-2-7B improves from 20.84\% to 28.51\% (+36.8\% relative), outperforming its baseline of 17.40\%. Llama-3.1-8B sees a moderate gain (62.91\% to 64.54\%, +2.6\%), while Viking-7B experiences a slight decline (29.25\% to 24.84\%, -15.1\%). 
For detailed per-language results on the SIB-200 benchmark, refer to \Cref{sec:detailed_results_sib200}

We intentionally omit FLORES-200 comparisons between bilingual and bilingual+code configurations because the fundamental language mixing issue identified in generation tasks as described in \Cref{sec:language_mixing} makes this comparison nonsensical.
As a reference, per-language BLEU scores are available in \Cref{sec:detailed_results_flores200}.

\subsection{SIB-200 Accuracy}
\label{sec:detailed_results_sib200}
The SIB-200 accuracy results are detailed across language categories: Table~\ref{tab:sib200_altruistic_scores} presents scores for altruistic languages, Table~\ref{tab:sib200_selfish_scores} for selfish languages, and Table~\ref{tab:sib200_stagnant_scores} for stagnant languages, covering various models and languages within each category.

\begin{table}[htbp]
\centering
\scriptsize
\renewcommand{\arraystretch}{1.2}
\resizebox{\textwidth}{!}{%
\begin{tabular}{l>{\columncolor[HTML]{DCE6F1}}c>{\columncolor[HTML]{DCE6F1}}c>{\columncolor[HTML]{DCE6F1}}c>{\columncolor[HTML]{DCE6F1}}c>{\columncolor[HTML]{DCE6F1}}c>{\columncolor[HTML]{DCE6F1}}ccccccccc}
\hline
\textbf{Model} & \textbf{zho\_Hans} & \textbf{mar\_Deva} & \textbf{ceb\_Latn} & \textbf{zul\_Latn} & \textbf{khm\_Khmr} & \textbf{eng\_Latn} & \textbf{hin\_Deva} & \textbf{tgl\_Latn} & \textbf{xho\_Latn} & \textbf{vie\_Latn} & \textbf{ilo\_Latn} & \textbf{npi\_Deva} & \textbf{yue\_Hant} & \textbf{ssw\_Latn} \\
\hline
L2-Bi-Alt & 0.2598 & 0.2108 & 0.2402 & 0.2794 & 0.1961 & 0.4216 & 0.1667 & 0.1912 & 0.2010 & 0.1961 & 0.1618 & 0.1716 & 0.2255 & 0.2010 \\
L2-Bi+Code-Alt & 0.3529 & 0.3284 & 0.4412 & 0.3627 & 0.3137 & 0.5049 & 0.2402 & 0.3873 & 0.3088 & 0.3382 & 0.3039 & 0.2451 & 0.3725 & 0.3137 \\
L2-Mono-Alt & 0.1765 & 0.1765 & 0.2108 & 0.1814 & 0.2010 & 0.2304 & 0.1471 & 0.1765 & 0.1667 & 0.2010 & 0.1618 & 0.1471 & 0.1863 & 0.1569 \\
L2-Mono+Code-Alt & 0.3529 & 0.3235 & 0.3922 & 0.3382 & 0.2598 & 0.4167 & 0.2598 & 0.3235 & 0.2843 & 0.3284 & 0.2990 & 0.2108 & 0.3186 & 0.3039 \\
Llama-2-7B (Base) & 0.3382 & 0.1765 & 0.2598 & 0.1569 & 0.1765 & 0.4020 & 0.2353 & 0.2647 & 0.1716 & 0.3088 & 0.2402 & 0.2402 & 0.3333 & 0.1618 \\
\hline
L3-Bi-Alt & 0.7500 & 0.6324 & 0.7010 & 0.6569 & 0.7059 & 0.7157 & 0.6127 & 0.6814 & 0.5637 & 0.6225 & 0.6471 & 0.5686 & 0.7647 & 0.5882 \\
L3-Bi+Code-Alt & 0.7157 & 0.6324 & 0.6765 & 0.6275 & 0.7010 & 0.7353 & 0.6422 & 0.6520 & 0.5049 & 0.6814 & 0.5931 & 0.5441 & 0.6912 & 0.5686 \\
L3-Mono-Alt & 0.6176 & 0.4510 & 0.4902 & 0.5196 & 0.4167 & 0.6176 & 0.3971 & 0.4265 & 0.4069 & 0.5343 & 0.3775 & 0.4020 & 0.6422 & 0.4461 \\
L3-Mono+Code-Alt & 0.6814 & 0.6324 & 0.6814 & 0.6765 & 0.5686 & 0.7843 & 0.5245 & 0.6127 & 0.5147 & 0.6961 & 0.5637 & 0.4804 & 0.7255 & 0.5784 \\
Llama-3.1-8B (Base) & 0.7549 & 0.6667 & 0.6912 & 0.5441 & 0.6422 & 0.7843 & 0.7010 & 0.7255 & 0.5392 & 0.7500 & 0.6765 & 0.6520 & 0.7647 & 0.4755 \\
\hline
V7-Bi-Alt & 0.2206 & 0.1814 & 0.2500 & 0.1618 & 0.1373 & 0.2500 & 0.1127 & 0.2353 & 0.1814 & 0.1422 & 0.1814 & 0.0931 & 0.1814 & 0.1569 \\
V7-Bi+Code-Alt & 0.3578 & 0.2206 & 0.2843 & 0.2451 & 0.2108 & 0.3137 & 0.1569 & 0.2451 & 0.2206 & 0.1569 & 0.2402 & 0.1716 & 0.3186 & 0.2010 \\
V7-Mono-Alt & 0.2157 & 0.1814 & 0.2353 & 0.1814 & 0.1716 & 0.2843 & 0.1618 & 0.1618 & 0.1618 & 0.2402 & 0.2157 & 0.1716 & 0.2206 & 0.1814 \\
V7-Mono+Code-Alt & 0.2157 & 0.1814 & 0.2990 & 0.2500 & 0.1912 & 0.2941 & 0.1569 & 0.2353 & 0.2108 & 0.1863 & 0.2255 & 0.1569 & 0.2451 & 0.2500 \\
Viking-7B (Base) & 0.3725 & 0.1814 & 0.2206 & 0.1471 & 0.1520 & 0.3235 & 0.1765 & 0.2010 & 0.1814 & 0.3186 & 0.2500 & 0.1961 & 0.4118 & 0.1520 \\
\hline
\end{tabular}
}
\caption{SIB-200 task accuracy for Altruistic languages across all models. Training language columns have a shaded background.}
\label{tab:sib200_altruistic_scores}
\end{table}

\begin{table}[htbp]
\centering
\scriptsize
\renewcommand{\arraystretch}{1.2}
\resizebox{\textwidth}{!}{%
\begin{tabular}{l>{\columncolor[HTML]{DCE6F1}}c>{\columncolor[HTML]{DCE6F1}}c>{\columncolor[HTML]{DCE6F1}}c>{\columncolor[HTML]{DCE6F1}}c>{\columncolor[HTML]{DCE6F1}}c>{\columncolor[HTML]{DCE6F1}}ccccccccccc}
\hline
\textbf{Model} & \textbf{deu\_Latn} & \textbf{bel\_Cyrl} & \textbf{mri\_Latn} & \textbf{kir\_Cyrl} & \textbf{nya\_Latn} & \textbf{eng\_Latn} & \textbf{fij\_Latn} & \textbf{bak\_Cyrl} & \textbf{dan\_Latn} & \textbf{rus\_Cyrl} & \textbf{smo\_Latn} & \textbf{bem\_Latn} & \textbf{kaz\_Cyrl} & \textbf{sna\_Latn} & \textbf{ukr\_Cyrl} & \textbf{nld\_Latn} \\
\hline
L2-Bi-Sel & 0.3088 & 0.3186 & 0.2500 & 0.1814 & 0.2353 & 0.4167 & 0.1618 & 0.1569 & 0.3578 & 0.3578 & 0.1765 & 0.1618 & 0.1569 & 0.1618 & 0.3235 & 0.4216 \\
L2-Bi+Code-Sel & 0.4265 & 0.3922 & 0.3186 & 0.2843 & 0.3333 & 0.5098 & 0.2206 & 0.2500 & 0.4461 & 0.4069 & 0.2206 & 0.2206 & 0.2794 & 0.2402 & 0.3922 & 0.4412 \\
L2-Mono-Sel & 0.4412 & 0.4020 & 0.2647 & 0.3775 & 0.2794 & 0.4461 & 0.1765 & 0.2549 & 0.4167 & 0.3725 & 0.1765 & 0.1912 & 0.2696 & 0.2010 & 0.4069 & 0.4216 \\
L2-Mono+Code-Sel & 0.4412 & 0.3775 & 0.3186 & 0.3186 & 0.2990 & 0.4412 & 0.2206 & 0.2843 & 0.3873 & 0.3676 & 0.2059 & 0.2451 & 0.2696 & 0.2255 & 0.3627 & 0.4216 \\
Llama-2-7B (Base) & 0.3922 & 0.2157 & 0.1912 & 0.1765 & 0.1765 & 0.4020 & 0.1765 & 0.1912 & 0.3627 & 0.2892 & 0.1765 & 0.1716 & 0.1667 & 0.1618 & 0.2941 & 0.3775 \\
\hline
L3-Bi-Sel & 0.7206 & 0.6078 & 0.5784 & 0.5735 & 0.6078 & 0.7451 & 0.3824 & 0.5294 & 0.6569 & 0.6373 & 0.3627 & 0.3627 & 0.5343 & 0.3627 & 0.6225 & 0.6373 \\
L3-Bi+Code-Sel & 0.6863 & 0.6127 & 0.6078 & 0.6127 & 0.6373 & 0.6716 & 0.4118 & 0.5294 & 0.6618 & 0.6569 & 0.3824 & 0.4461 & 0.6029 & 0.3676 & 0.6029 & 0.6716 \\
L3-Mono-Sel & 0.7059 & 0.5833 & 0.5980 & 0.5833 & 0.5784 & 0.6520 & 0.3971 & 0.5147 & 0.6618 & 0.5637 & 0.4363 & 0.4167 & 0.5686 & 0.3775 & 0.5196 & 0.6127 \\
L3-Mono+Code-Sel & 0.7451 & 0.6863 & 0.6471 & 0.7108 & 0.6471 & 0.7108 & 0.3775 & 0.5784 & 0.7402 & 0.7010 & 0.3627 & 0.4069 & 0.6618 & 0.3725 & 0.7304 & 0.7059 \\
Llama-3.1-8B (Base) & 0.7598 & 0.7206 & 0.6029 & 0.7157 & 0.5539 & 0.7843 & 0.4559 & 0.6961 & 0.7451 & 0.7157 & 0.5931 & 0.4559 & 0.7304 & 0.4706 & 0.7402 & 0.7206 \\
\hline
V7-Bi-Sel & 0.3088 & 0.2745 & 0.2255 & 0.2549 & 0.3333 & 0.3578 & 0.2255 & 0.2157 & 0.2990 & 0.2598 & 0.1961 & 0.2108 & 0.2157 & 0.2010 & 0.2451 & 0.2990 \\
V7-Bi+Code-Sel & 0.3039 & 0.3529 & 0.2206 & 0.2451 & 0.2549 & 0.3676 & 0.1716 & 0.1961 & 0.3039 & 0.3039 & 0.1716 & 0.1814 & 0.2010 & 0.2010 & 0.2549 & 0.2206 \\
V7-Mono-Sel & 0.2549 & 0.3627 & 0.1814 & 0.2059 & 0.2353 & 0.2745 & 0.1520 & 0.1814 & 0.2108 & 0.2059 & 0.1667 & 0.1520 & 0.1618 & 0.1520 & 0.1912 & 0.1814 \\
V7-Mono+Code-Sel & 0.3431 & 0.3676 & 0.3578 & 0.3578 & 0.3775 & 0.3922 & 0.2402 & 0.2500 & 0.3873 & 0.3382 & 0.2206 & 0.2598 & 0.2745 & 0.2451 & 0.3186 & 0.3186 \\
Viking-7B (Base) & 0.3480 & 0.2843 & 0.1667 & 0.2059 & 0.1667 & 0.3235 & 0.1961 & 0.2451 & 0.3627 & 0.3529 & 0.1961 & 0.1569 & 0.2108 & 0.1618 & 0.3529 & 0.4020 \\
\hline
\end{tabular}
}
\caption{SIB-200 task accuracy for Selfish languages across all models. Training language columns have a shaded background.}
\label{tab:sib200_selfish_scores}
\end{table}

\begin{table}[htbp]
\centering
\renewcommand{\arraystretch}{1.2}
\resizebox{\textwidth}{!}{%
\begin{tabular}{l>{\columncolor[HTML]{DCE6F1}}c>{\columncolor[HTML]{DCE6F1}}c>{\columncolor[HTML]{DCE6F1}}c>{\columncolor[HTML]{DCE6F1}}cccccccc}
\hline
\textbf{Model} & \textbf{tha\_Thai} & \textbf{yor\_Latn} & \textbf{sna\_Latn} & \textbf{wol\_Latn} & \textbf{nya\_Latn} & \textbf{zul\_Latn} & \textbf{shn\_Mymr} & \textbf{bam\_Latn} & \textbf{hau\_Latn} & \textbf{ibo\_Latn} & \textbf{lao\_Laoo} \\
\hline
L2-Bi-Stag & 0.2598 & 0.1765 & 0.1961 & 0.1618 & 0.1569 & 0.1471 & 0.1569 & 0.1471 & 0.1471 & 0.1520 & 0.1520 \\
L2-Bi+Code-Stag & 0.3186 & 0.2108 & 0.2255 & 0.1912 & 0.1912 & 0.1814 & 0.1863 & 0.1618 & 0.1618 & 0.2010 & 0.1667 \\
L2-Mono-Stag & 0.3137 & 0.2402 & 0.2549 & 0.1765 & 0.1618 & 0.1618 & 0.1471 & 0.1618 & 0.1520 & 0.1569 & 0.1373 \\
L2-Mono+Code-Stag & 0.3480 & 0.3039 & 0.3529 & 0.2255 & 0.2255 & 0.1961 & 0.1814 & 0.1912 & 0.1961 & 0.2010 & 0.1569 \\
Llama-2-7B (Base) & 0.2353 & 0.1569 & 0.1618 & 0.1961 & 0.1765 & 0.1569 & 0.1863 & 0.1667 & 0.1667 & 0.1667 & 0.1569 \\
\hline
L3-Bi-Stag & 0.7157 & 0.6078 & 0.6667 & 0.5637 & 0.5441 & 0.3971 & 0.3382 & 0.3431 & 0.3382 & 0.4020 & 0.3775 \\
L3-Bi+Code-Stag & 0.7696 & 0.6471 & 0.6520 & 0.6422 & 0.5490 & 0.4069 & 0.3284 & 0.3529 & 0.4118 & 0.4118 & 0.2941 \\
L3-Mono-Stag & 0.5784 & 0.4510 & 0.5539 & 0.4706 & 0.3431 & 0.3480 & 0.3137 & 0.3039 & 0.3235 & 0.2696 & 0.2598 \\
L3-Mono+Code-Stag & 0.5637 & 0.5588 & 0.5882 & 0.5147 & 0.4167 & 0.3922 & 0.2990 & 0.3333 & 0.3480 & 0.3676 & 0.2549 \\
Llama-3.1-8B (Base) & 0.7451 & 0.5245 & 0.4706 & 0.4853 & 0.5539 & 0.5441 & 0.4657 & 0.3971 & 0.6716 & 0.6520 & 0.5441 \\
\hline
V7-Bi-Stag & 0.4412 & 0.4118 & 0.4461 & 0.4216 & 0.2647 & 0.2206 & 0.1569 & 0.2647 & 0.2255 & 0.2059 & 0.1667 \\
V7-Bi+Code-Stag & 0.2647 & 0.2745 & 0.2745 & 0.2598 & 0.1569 & 0.1471 & 0.1225 & 0.1667 & 0.1422 & 0.1176 & 0.1078 \\
V7-Mono-Stag & 0.2549 & 0.2255 & 0.2598 & 0.2255 & 0.2010 & 0.1912 & 0.1422 & 0.1912 & 0.1814 & 0.1765 & 0.0980 \\
V7-Mono+Code-Stag & 0.3480 & 0.2794 & 0.3284 & 0.2157 & 0.2206 & 0.1814 & 0.2059 & 0.2304 & 0.1814 & 0.1912 & 0.1863 \\
Viking-7B (Base) & 0.3725 & 0.2108 & 0.1618 & 0.2206 & 0.1667 & 0.1471 & 0.1863 & 0.1667 & 0.1569 & 0.1667 & 0.2010 \\
\hline
\end{tabular}
}
\caption{SIB-200 task accuracy for Stagnant languages across all models. Training language columns have a shaded background.}
\label{tab:sib200_stagnant_scores}
\end{table}

\subsection{FLORES-200 BLEU Scores}
\label{sec:detailed_results_flores200}
The BLEU scores for the FLORES-200 benchmark are detailed across language categories and translation directions: Tables~\ref{tab:flores200_altruistic_engx} and~\ref{tab:flores200_altruistic_xeng} present scores for altruistic languages (Eng-X and X-Eng, respectively), Tables~\ref{tab:flores200_selfish_engx} and~\ref{tab:flores200_selfish_xeng} for selfish languages (Eng-X and X-Eng), and Tables~\ref{tab:flores200_stagnant_engx} and~\ref{tab:flores200_stagnant_xeng} for stagnant languages (Eng-X and X-Eng)

\begin{table}[htbp]
\centering
\scriptsize
\renewcommand{\arraystretch}{1.2}
\resizebox{\textwidth}{!}{%
\begin{tabular}{lccccccccccccccc}
\hline
\textbf{Language Pair} & \textbf{L2-Bi-Alt} & \textbf{L2-Bi+Code-Alt} & \textbf{L2-Mono-Alt} & \textbf{L2-Mono+Code-Alt} & \textbf{Llama-2-7B} & \textbf{L3-Bi-Alt} & \textbf{L3-Bi+Code-Alt} & \textbf{L3-Mono-Alt} & \textbf{L3-Mono+Code-Alt} & \textbf{Llama-3.1-8B} & \textbf{V7-Bi-Alt} & \textbf{V7-Bi+Code-Alt} & \textbf{V7-Mono-Alt} & \textbf{V7-Mono+Code-Alt} & \textbf{Viking-7B} \\
\hline
\rowcolor[HTML]{DCE6F1} eng\_Latn-zho\_Hans & 9.62 & 4.10 & 10.23 & 10.13 & 10.47 & 2.87 & 5.53 & 17.14 & 17.34 & 24.27 & 0.80 & 2.07 & 1.18 & 2.10 & 9.72 \\
\rowcolor[HTML]{DCE6F1} eng\_Latn-ceb\_Latn & 19.37 & 3.59 & 19.46 & 19.63 & 5.35 & 0.75 & 1.51 & 20.81 & 20.37 & 22.72 & 1.95 & 3.50 & 6.27 & 6.88 & 3.66 \\
\rowcolor[HTML]{DCE6F1} eng\_Latn-mar\_Deva & 8.44 & 14.81 & 8.93 & 8.63 & 1.39 & 4.22 & 5.45 & 9.20 & 8.33 & 6.83 & 6.21 & 7.24 & 0.86 & 1.05 & 0.21 \\
\rowcolor[HTML]{DCE6F1} eng\_Latn-zul\_Latn & 6.56 & 8.31 & 6.54 & 6.54 & 1.64 & 6.22 & 6.77 & 9.59 & 9.70 & 26.17 & 12.55 & 12.62 & 1.63 & 2.07 & 0.94 \\
\rowcolor[HTML]{DCE6F1} eng\_Latn-khm\_Khmr & 3.03 & 2.84 & 3.27 & 3.38 & 0.09 & 4.69 & 5.02 & 8.46 & 8.30 & 1.76 & 4.13 & 4.19 & 1.59 & 1.54 & 0.07 \\
eng\_Latn-npi\_Deva & 1.40 & 2.01 & 1.41 & 1.49 & 1.53 & 0.66 & 0.93 & 1.35 & 1.29 & 6.13 & 0.80 & 0.99 & 0.07 & 0.08 & 0.28 \\
eng\_Latn-vie\_Latn & 6.15 & 0.71 & 6.83 & 6.47 & 15.44 & 0.70 & 0.79 & 13.23 & 16.16 & 26.63 & 0.55 & 1.03 & 0.09 & 0.29 & 5.30 \\
eng\_Latn-tgl\_Latn & 5.81 & 1.62 & 5.79 & 6.25 & 7.32 & 1.23 & 1.43 & 5.67 & 5.67 & 15.14 & 0.98 & 1.83 & 1.32 & 2.06 & 4.23 \\
eng\_Latn-ssw\_Latn & 3.34 & 3.72 & 3.27 & 3.61 & 1.54 & 2.78 & 2.72 & 3.93 & 4.08 & 3.04 & 4.29 & 4.41 & 0.90 & 0.79 & 0.82 \\
eng\_Latn-xho\_Latn & 3.86 & 3.71 & 3.44 & 4.02 & 1.91 & 2.83 & 2.96 & 4.39 & 4.64 & 3.55 & 5.63 & 5.44 & 1.14 & 1.01 & 1.13 \\
eng\_Latn-yue\_Hant & 6.81 & 1.39 & 8.51 & 7.59 & 8.15 & 1.26 & 2.80 & 14.58 & 14.54 & 4.63 & 0.37 & 1.40 & 0.51 & 0.88 & 6.50 \\
eng\_Latn-ilo\_Latn & 3.55 & 1.30 & 3.58 & 3.65 & 2.97 & 0.79 & 1.01 & 3.48 & 3.48 & 25.82 & 0.82 & 1.34 & 0.77 & 0.78 & 2.34 \\
eng\_Latn-hin\_Deva & 2.09 & 3.14 & 1.96 & 1.79 & 5.27 & 1.05 & 1.42 & 3.17 & 2.80 & 24.30 & 1.44 & 1.48 & 0.14 & 0.18 & 1.29 \\
\hline
\end{tabular}
}
\caption{FLORES-200 BLEU scores for Altruistic languages (Eng-X). Training language rows have a shaded background.}
\label{tab:flores200_altruistic_engx}
\end{table}

\begin{table}[htbp]
\centering
\scriptsize
\renewcommand{\arraystretch}{1.2}
\resizebox{\textwidth}{!}{%
\begin{tabular}{lccccccccccccccc}
\hline
\textbf{Language Pair} & \textbf{L2-Bi-Alt} & \textbf{L2-Bi+Code-Alt} & \textbf{L2-Mono-Alt} & \textbf{L2-Mono+Code-Alt} & \textbf{Llama-2-7B} & \textbf{L3-Bi-Alt} & \textbf{L3-Bi+Code-Alt} & \textbf{L3-Mono-Alt} & \textbf{L3-Mono+Code-Alt} & \textbf{Llama-3.1-8B} & \textbf{V7-Bi-Alt} & \textbf{V7-Bi+Code-Alt} & \textbf{V7-Mono-Alt} & \textbf{V7-Mono+Code-Alt} & \textbf{Viking-7B} \\
\hline
\rowcolor[HTML]{DCE6F1} zho\_Hans-eng\_Latn & 18.35 & 13.03 & 16.85 & 17.97 & 18.28 & 6.62 & 9.40 & 18.99 & 19.68 & 22.43 & 0.86 & 4.94 & 2.47 & 3.02 & 16.06 \\
\rowcolor[HTML]{DCE6F1} ceb\_Latn-eng\_Latn & 29.85 & 11.16 & 29.36 & 29.81 & 9.58 & 8.12 & 10.83 & 29.98 & 28.10 & 22.67 & 8.52 & 14.92 & 6.74 & 9.15 & 6.03 \\
\rowcolor[HTML]{DCE6F1} mar\_Deva-eng\_Latn & 17.12 & 5.35 & 16.63 & 17.59 & 4.09 & 1.79 & 3.67 & 19.52 & 19.69 & 22.38 & 0.14 & 0.24 & 0.99 & 0.91 & 0.55 \\
\rowcolor[HTML]{DCE6F1} zul\_Latn-eng\_Latn & 19.04 & 8.26 & 18.28 & 18.81 & 3.05 & 2.67 & 4.39 & 20.72 & 20.77 & 8.73 & 0.12 & 0.78 & 3.10 & 4.07 & 2.33 \\
\rowcolor[HTML]{DCE6F1} vie\_Latn-eng\_Latn & 20.99 & 10.85 & 19.78 & 19.97 & 20.61 & 8.57 & 9.11 & 21.97 & 22.70 & 26.12 & 0.08 & 0.19 & 0.13 & 0.40 & 10.32 \\
\rowcolor[HTML]{DCE6F1} khm\_Khmr-eng\_Latn & 13.49 & 1.64 & 12.77 & 13.10 & 2.06 & 0.76 & 2.41 & 16.49 & 17.67 & 15.51 & 0.22 & 0.67 & 0.79 & 1.01 & 0.81 \\
ssw\_Latn-eng\_Latn & 8.47 & 3.82 & 8.04 & 8.80 & 3.16 & 1.87 & 1.94 & 8.78 & 8.88 & 6.29 & 0.13 & 0.55 & 1.31 & 1.99 & 2.18 \\
npi\_Deva-eng\_Latn & 3.25 & 0.94 & 2.70 & 3.29 & 4.69 & 1.18 & 1.70 & 6.40 & 7.10 & 22.81 & 0.04 & 0.20 & 0.18 & 0.26 & 0.75 \\
yue\_Hant-eng\_Latn & 17.55 & 8.00 & 16.45 & 17.63 & 18.66 & 4.52 & 6.94 & 18.94 & 19.45 & 23.26 & 0.31 & 2.63 & 1.78 & 2.91 & 14.27 \\
tgl\_Latn-eng\_Latn & 13.83 & 7.81 & 13.95 & 14.52 & 16.29 & 4.67 & 6.07 & 15.10 & 14.91 & 28.92 & 0.38 & 2.33 & 1.57 & 2.15 & 6.74 \\
hin\_Deva-eng\_Latn & 6.62 & 2.13 & 6.77 & 7.72 & 12.10 & 2.33 & 3.80 & 16.14 & 16.38 & 27.20 & 0.05 & 0.11 & 0.31 & 0.21 & 1.04 \\
ilo\_Latn-eng\_Latn & 5.54 & 2.24 & 5.34 & 5.28 & 5.67 & 1.23 & 1.77 & 6.06 & 5.16 & 15.19 & 0.19 & 0.62 & 0.60 & 0.95 & 4.23 \\
xho\_Latn-eng\_Latn & 9.56 & 4.88 & 9.00 & 9.70 & 3.35 & 1.98 & 2.83 & 8.88 & 9.75 & 8.83 & 0.17 & 0.87 & 1.65 & 2.62 & 2.62 \\
\hline
\end{tabular}
}
\caption{FLORES-200 BLEU scores for Altruistic languages (X-Eng). Training language rows have a shaded background.}
\label{tab:flores200_altruistic_xeng}
\end{table}

\begin{table}[htbp]
\centering
\scriptsize
\renewcommand{\arraystretch}{1.2}
\resizebox{\textwidth}{!}{%
\begin{tabular}{lccccccccccccccc}
\hline
\textbf{Language Pair} & \textbf{L2-Bi+Code-Sel} & \textbf{L2-Bi-Sel} & \textbf{L2-Mono+Code-Sel} & \textbf{L2-Mono-Sel} & \textbf{Llama-2-7B} & \textbf{L3-Bi+Code-Sel} & \textbf{L3-Bi-Sel} & \textbf{L3-Mono+Code-Sel} & \textbf{L3-Mono-Sel} & \textbf{Llama-3.1-8B} & \textbf{V7-Bi+Code-Sel} & \textbf{V7-Bi-Sel} & \textbf{V7-Mono+Code-Sel} & \textbf{V7-Mono-Sel} & \textbf{Viking-7B} \\
\hline
\rowcolor[HTML]{DCE6F1} eng\_Latn-deu\_Latn & 18.51 & 8.50 & 23.16 & 22.85 & 23.96 & 11.00 & 8.43 & 22.19 & 24.78 & 27.08 & 16.69 & 11.15 & 12.22 & 6.09 & 20.45 \\
\rowcolor[HTML]{DCE6F1} eng\_Latn-bel\_Cyrl & 2.63 & 1.65 & 12.27 & 11.81 & 1.95 & 3.26 & 0.82 & 11.98 & 14.12 & 11.23 & 0.59 & 0.24 & 4.00 & 0.82 & 0.98 \\
\rowcolor[HTML]{DCE6F1} eng\_Latn-mri\_Latn & 7.13 & 3.60 & 4.88 & 3.94 & 2.50 & 3.88 & 2.88 & 5.07 & 6.15 & 4.55 & 6.43 & 4.92 & 1.05 & 0.50 & 0.83 \\
\rowcolor[HTML]{DCE6F1} eng\_Latn-kir\_Cyrl & 4.60 & 2.73 & 4.01 & 3.76 & 1.71 & 3.60 & 2.72 & 6.51 & 7.09 & 0.90 & 3.18 & 1.53 & 1.26 & 0.39 & 0.78 \\
\rowcolor[HTML]{DCE6F1} eng\_Latn-nya\_Latn & 4.40 & 3.34 & 6.76 & 6.30 & 1.65 & 4.65 & 3.22 & 6.44 & 8.06 & 2.98 & 8.59 & 7.84 & 1.59 & 0.51 & 0.86 \\
eng\_Latn-sna\_Latn & 0.92 & 0.61 & 1.11 & 1.03 & 1.73 & 0.92 & 0.65 & 1.45 & 1.56 & 3.67 & 1.24 & 1.15 & 0.25 & 0.11 & 0.94 \\
eng\_Latn-bak\_Cyrl & 1.29 & 0.63 & 1.48 & 1.31 & 1.67 & 1.09 & 0.78 & 2.57 & 2.55 & 7.11 & 0.98 & 0.49 & 0.45 & 0.31 & 0.60 \\
eng\_Latn-nld\_Latn & 9.08 & 2.24 & 15.86 & 13.76 & 18.00 & 3.37 & 1.07 & 14.38 & 11.25 & 20.31 & 1.87 & 0.99 & 1.87 & 0.68 & 16.44 \\
eng\_Latn-kaz\_Cyrl & 1.36 & 0.63 & 1.70 & 1.52 & 1.54 & 1.18 & 0.79 & 2.90 & 3.02 & 6.93 & 1.08 & 0.62 & 0.65 & 0.36 & 0.79 \\
eng\_Latn-fij\_Latn & 1.05 & 0.63 & 0.91 & 0.65 & 1.75 & 0.90 & 0.64 & 0.83 & 0.96 & 3.32 & 1.31 & 0.89 & 0.23 & 0.09 & 1.32 \\
eng\_Latn-smo\_Latn & 1.66 & 0.88 & 1.04 & 0.66 & 1.76 & 1.00 & 0.85 & 1.01 & 0.90 & 11.34 & 1.16 & 0.98 & 0.18 & 0.07 & 1.09 \\
eng\_Latn-rus\_Cyrl & 2.13 & 0.76 & 13.87 & 12.88 & 21.99 & 2.56 & 1.01 & 16.71 & 16.58 & 4.01 & 1.24 & 0.38 & 1.94 & 0.44 & 11.78 \\
eng\_Latn-dan\_Latn & 7.51 & 2.55 & 18.75 & 16.45 & 21.74 & 4.31 & 1.24 & 16.89 & 15.19 & 1.37 & 3.05 & 0.80 & 2.85 & 1.05 & 38.18 \\
eng\_Latn-ukr\_Cyrl & 0.59 & 0.45 & 2.23 & 2.19 & 18.59 & 0.61 & 0.35 & 3.45 & 3.23 & 7.14 & 0.41 & 0.15 & 0.40 & 0.09 & 8.87 \\
eng\_Latn-bem\_Latn & 1.48 & 0.91 & 1.00 & 0.86 & 1.34 & 1.25 & 0.95 & 1.31 & 1.54 & 14.91 & 1.13 & 1.26 & 0.53 & 0.19 & 0.48 \\
\hline
\end{tabular}
}
\caption{FLORES-200 BLEU scores for Selfish languages (Eng-X). Training language rows have a shaded background.}
\label{tab:flores200_selfish_engx}
\end{table}

\begin{table}[htbp]
\centering
\scriptsize
\renewcommand{\arraystretch}{1.2}
\resizebox{\textwidth}{!}{%
\begin{tabular}{lccccccccccccccc}
\hline
\textbf{Language Pair} & \textbf{L2-Bi+Code-Sel} & \textbf{L2-Bi-Sel} & \textbf{L2-Mono+Code-Sel} & \textbf{L2-Mono-Sel} & \textbf{Llama-2-7B} & \textbf{L3-Bi+Code-Sel} & \textbf{L3-Bi-Sel} & \textbf{L3-Mono+Code-Sel} & \textbf{L3-Mono-Sel} & \textbf{Llama-3.1-8B} & \textbf{V7-Bi+Code-Sel} & \textbf{V7-Bi-Sel} & \textbf{V7-Mono+Code-Sel} & \textbf{V7-Mono-Sel} & \textbf{Viking-7B} \\
\hline
\rowcolor[HTML]{DCE6F1} deu\_Latn-eng\_Latn & 29.31 & 9.88 & 32.13 & 32.34 & 27.87 & 12.85 & 8.32 & 31.02 & 32.05 & 33.51 & 10.51 & 2.28 & 20.36 & 15.89 & 31.29 \\
\rowcolor[HTML]{DCE6F1} bel\_Cyrl-eng\_Latn & 15.59 & 5.61 & 18.95 & 18.44 & 8.78 & 7.54 & 5.36 & 16.67 & 18.10 & 19.36 & 5.43 & 4.89 & 7.73 & 7.02 & 3.49 \\
\rowcolor[HTML]{DCE6F1} mri\_Latn-eng\_Latn & 2.69 & 0.72 & 10.66 & 10.40 & 4.21 & 1.55 & 0.08 & 10.88 & 12.22 & 11.15 & 0.33 & 0.06 & 3.20 & 1.71 & 1.86 \\
\rowcolor[HTML]{DCE6F1} kir\_Cyrl-eng\_Latn & 4.15 & 0.99 & 10.43 & 10.52 & 3.29 & 2.81 & 0.19 & 13.02 & 13.63 & 14.98 & 0.86 & 0.23 & 2.88 & 2.32 & 1.93 \\
\rowcolor[HTML]{DCE6F1} nya\_Latn-eng\_Latn & 1.31 & 0.20 & 15.84 & 16.07 & 2.66 & 2.11 & 0.12 & 15.48 & 17.25 & 6.54 & 0.39 & 0.10 & 4.92 & 3.03 & 2.43 \\
ukr\_Cyrl-eng\_Latn & 23.49 & 7.43 & 26.05 & 26.35 & 26.16 & 8.75 & 7.09 & 24.64 & 25.77 & 30.98 & 2.54 & 0.72 & 4.72 & 4.09 & 24.78 \\
nld\_Latn-eng\_Latn & 21.81 & 6.73 & 24.14 & 24.52 & 20.21 & 7.64 & 4.94 & 21.27 & 22.88 & 24.35 & 1.47 & 0.35 & 5.53 & 3.37 & 22.61 \\
dan\_Latn-eng\_Latn & 31.12 & 10.03 & 34.53 & 34.89 & 29.78 & 10.41 & 6.87 & 30.69 & 31.68 & 35.30 & 12.17 & 3.50 & 24.14 & 18.36 & 39.68 \\
rus\_Cyrl-eng\_Latn & 23.10 & 7.64 & 26.38 & 26.13 & 23.66 & 9.04 & 7.11 & 23.96 & 25.05 & 27.08 & 5.98 & 1.26 & 9.05 & 7.87 & 23.83 \\
smo\_Latn-eng\_Latn & 0.95 & 0.38 & 3.02 & 3.29 & 2.92 & 0.70 & 0.06 & 2.88 & 3.10 & 9.34 & 0.10 & 0.05 & 0.97 & 0.43 & 1.78 \\
bak\_Cyrl-eng\_Latn & 1.55 & 0.53 & 3.66 & 3.86 & 4.07 & 1.38 & 0.11 & 7.26 & 6.97 & 18.59 & 0.56 & 0.18 & 1.03 & 0.71 & 1.69 \\
fij\_Latn-eng\_Latn & 0.74 & 0.21 & 2.02 & 1.94 & 2.53 & 0.38 & 0.03 & 2.20 & 1.90 & 4.52 & 0.12 & 0.09 & 0.71 & 0.55 & 2.07 \\
kaz\_Cyrl-eng\_Latn & 1.87 & 0.62 & 4.61 & 4.34 & 3.64 & 1.89 & 0.19 & 8.53 & 9.46 & 20.01 & 0.56 & 0.16 & 1.44 & 0.88 & 2.24 \\
sna\_Latn-eng\_Latn & 0.68 & 0.32 & 3.77 & 3.40 & 2.90 & 0.83 & 0.09 & 3.27 & 3.47 & 7.09 & 0.31 & 0.07 & 1.40 & 0.73 & 2.50 \\
bem\_Latn-eng\_Latn & 0.81 & 0.26 & 3.74 & 3.36 & 2.73 & 0.60 & 0.10 & 3.04 & 2.84 & 4.89 & 0.12 & 0.04 & 1.30 & 0.93 & 2.42 \\
\hline
\end{tabular}
}
\caption{FLORES-200 BLEU scores for Selfish languages (X-Eng). Training language rows have a shaded background.}
\label{tab:flores200_selfish_xeng}
\end{table}

\begin{table}[htbp]
\centering
\scriptsize
\renewcommand{\arraystretch}{1.2}
\resizebox{\textwidth}{!}{%
\begin{tabular}{lccccccccccccccc}
\hline
\textbf{Language Pair} & \textbf{L2-Bi+Code-Stag} & \textbf{L2-Bi-Stag} & \textbf{L2-Mono+Code-Stag} & \textbf{L2-Mono-Stag} & \textbf{Llama-2-7B} & \textbf{L3-Bi+Code-Stag} & \textbf{L3-Bi-Stag} & \textbf{L3-Mono+Code-Stag} & \textbf{L3-Mono-Stag} & \textbf{Llama-3.1-8B} & \textbf{V7-Bi+Code-Stag} & \textbf{V7-Bi-Stag} & \textbf{V7-Mono+Code-Stag} & \textbf{V7-Mono-Stag} & \textbf{Viking-7B} \\
\hline
\rowcolor[HTML]{DCE6F1} eng\_Latn-tha\_Thai & 23.11 & 21.99 & 18.06 & 16.84 & 3.60 & 10.11 & 8.48 & 20.76 & 21.85 & 19.44 & 15.05 & 16.23 & 4.02 & 3.27 & 2.98 \\
\rowcolor[HTML]{DCE6F1} eng\_Latn-yor\_Latn & 1.29 & 1.15 & 1.84 & 1.96 & 0.55 & 1.08 & 0.90 & 2.59 & 2.57 & 2.69 & 2.37 & 2.29 & 0.69 & 0.67 & 0.60 \\
\rowcolor[HTML]{DCE6F1} eng\_Latn-sna\_Latn & 4.07 & 3.27 & 5.07 & 4.83 & 1.73 & 4.62 & 3.55 & 6.74 & 7.21 & 3.67 & 8.67 & 10.10 & 1.37 & 1.52 & 0.94 \\
\rowcolor[HTML]{DCE6F1} eng\_Latn-wol\_Latn & 0.29 & 0.30 & 1.05 & 0.96 & 0.97 & 0.38 & 0.25 & 1.12 & 1.25 & 2.24 & 0.58 & 0.55 & 0.24 & 0.25 & 0.85 \\
eng\_Latn-hau\_Latn & 0.44 & 0.52 & 0.68 & 0.66 & 0.73 & 0.72 & 0.54 & 1.31 & 1.40 & 6.63 & 0.26 & 0.43 & 0.21 & 0.32 & 0.87 \\
eng\_Latn-shn\_Mymr & 0.25 & 0.20 & 0.06 & 0.11 & 0.00 & 0.26 & 0.15 & 0.12 & 0.01 & 0.28 & 0.11 & 0.26 & 0.07 & 0.08 & 0.03 \\
eng\_Latn-nya\_Latn & 0.71 & 0.59 & 1.30 & 1.52 & 1.65 & 0.65 & 0.59 & 1.67 & 1.88 & 2.98 & 0.83 & 0.94 & 0.78 & 0.56 & 0.86 \\
eng\_Latn-zul\_Latn & 0.75 & 0.69 & 1.53 & 1.54 & 1.64 & 0.79 & 0.69 & 1.89 & 2.13 & 26.17 & 0.97 & 1.35 & 0.45 & 0.42 & 0.94 \\
eng\_Latn-lao\_Laoo & 0.25 & 0.32 & 0.16 & 0.24 & 0.05 & 0.18 & 0.13 & 0.18 & 0.31 & 3.68 & 0.22 & 0.37 & 0.19 & 0.06 & 0.09 \\
eng\_Latn-ibo\_Latn & 0.77 & 0.64 & 0.71 & 0.80 & 0.56 & 0.64 & 0.54 & 1.14 & 1.26 & 5.45 & 0.56 & 0.63 & 0.12 & 0.18 & 0.59 \\
eng\_Latn-bam\_Latn & 0.13 & 0.12 & 0.61 & 0.55 & 0.53 & 0.31 & 0.08 & 0.59 & 0.63 & 22.51 & 0.11 & 0.48 & 0.21 & 0.17 & 0.20 \\
\hline
\end{tabular}
}
\caption{FLORES-200 BLEU scores for Stagnant languages (Eng-X). Training language rows have a shaded background.}
\label{tab:flores200_stagnant_engx}
\end{table}

\begin{table}[htbp]
\centering
\scriptsize
\renewcommand{\arraystretch}{1.2}
\resizebox{\textwidth}{!}{%
\begin{tabular}{lccccccccccccccc}
\hline
\textbf{Language Pair} & \textbf{L2-Bi+Code-Stag} & \textbf{L2-Bi-Stag} & \textbf{L2-Mono+Code-Stag} & \textbf{L2-Mono-Stag} & \textbf{Llama-2-7B} & \textbf{L3-Bi+Code-Stag} & \textbf{L3-Bi-Stag} & \textbf{L3-Mono+Code-Stag} & \textbf{L3-Mono-Stag} & \textbf{Llama-3.1-8B} & \textbf{V7-Bi+Code-Stag} & \textbf{V7-Bi-Stag} & \textbf{V7-Mono+Code-Stag} & \textbf{V7-Mono-Stag} & \textbf{Viking-7B} \\
\hline
\rowcolor[HTML]{DCE6F1} tha\_Thai-eng\_Latn & 1.744 & 0.491 & 17.486 & 16.364 & 5.85 & 1.944 & 0.062 & 21.167 & 21.349 & 22.72 & 0.112 & 0.061 & 2.501 & 3.396 & 3.15 \\
\rowcolor[HTML]{DCE6F1} yor\_Latn-eng\_Latn & 0.181 & 0.049 & 8.495 & 8.500 & 2.08 & 0.359 & 0.026 & 9.224 & 10.366 & 6.48 & 0.065 & 0.042 & 1.761 & 1.647 & 1.54 \\
\rowcolor[HTML]{DCE6F1} sna\_Latn-eng\_Latn & 0.245 & 0.016 & 13.943 & 13.119 & 2.9 & 1.268 & 0.307 & 15.935 & 17.034 & 7.09 & 0.061 & 0.064 & 3.345 & 3.766 & 2.5 \\
\rowcolor[HTML]{DCE6F1} wol\_Latn-eng\_Latn & 0.091 & 0.040 & 4.723 & 4.372 & 2.91 & 0.514 & 0.191 & 6.461 & 6.521 & 4.69 & 0.041 & 0.039 & 0.832 & 0.842 & 2.4 \\
hau\_Latn-eng\_Latn & 0.128 & 0.037 & 2.024 & 1.949 & 2.25 & 0.281 & 0.177 & 2.324 & 2.083 & 14.55 & 0.026 & 0.030 & 0.441 & 0.366 & 1.75 \\
bam\_Latn-eng\_Latn & 0.077 & 0.034 & 2.425 & 2.314 & 2.11 & 0.215 & 0.078 & 2.145 & 2.105 & 3.38 & 0.054 & 0.028 & 0.255 & 0.401 & 1.97 \\
shn\_Mymr-eng\_Latn & 0.127 & 0.073 & 2.494 & 2.072 & 1.96 & 0.433 & 0.053 & 2.433 & 1.753 & 5.35 & 0.073 & 0.076 & 0.238 & 0.143 & 0.88 \\
nya\_Latn-eng\_Latn & 0.223 & 0.071 & 3.048 & 3.454 & 2.66 & 0.535 & 0.269 & 3.470 & 3.332 & 6.54 & 0.049 & 0.042 & 0.614 & 0.585 & 2.43 \\
zul\_Latn-eng\_Latn & 0.180 & 0.065 & 2.913 & 2.980 & 3.05 & 0.381 & 0.201 & 3.165 & 2.960 & 8.73 & 0.046 & 0.041 & 0.337 & 0.480 & 2.33 \\
lao\_Laoo-eng\_Latn & 0.104 & 0.061 & 2.142 & 1.834 & 2.06 & 0.394 & 0.053 & 2.080 & 1.993 & 9.88 & 0.050 & 0.045 & 0.405 & 0.258 & 1.37 \\
ibo\_Latn-eng\_Latn & 0.218 & 0.034 & 2.158 & 2.192 & 2.23 & 0.359 & 0.126 & 2.402 & 2.066 & 12.3 & 0.037 & 0.026 & 0.346 & 0.263 & 1.48 \\
\hline
\end{tabular}
}
\caption{FLORES-200 BLEU scores for Stagnant languages (X-Eng). Training language rows have a shaded background.}
\label{tab:flores200_stagnant_xeng}
\end{table}

\subsection{Language Mixing Examples}
\label{sec:more_language_mixing}
\begin{figure}[htbp]
\centering
\includegraphics[width=\textwidth]{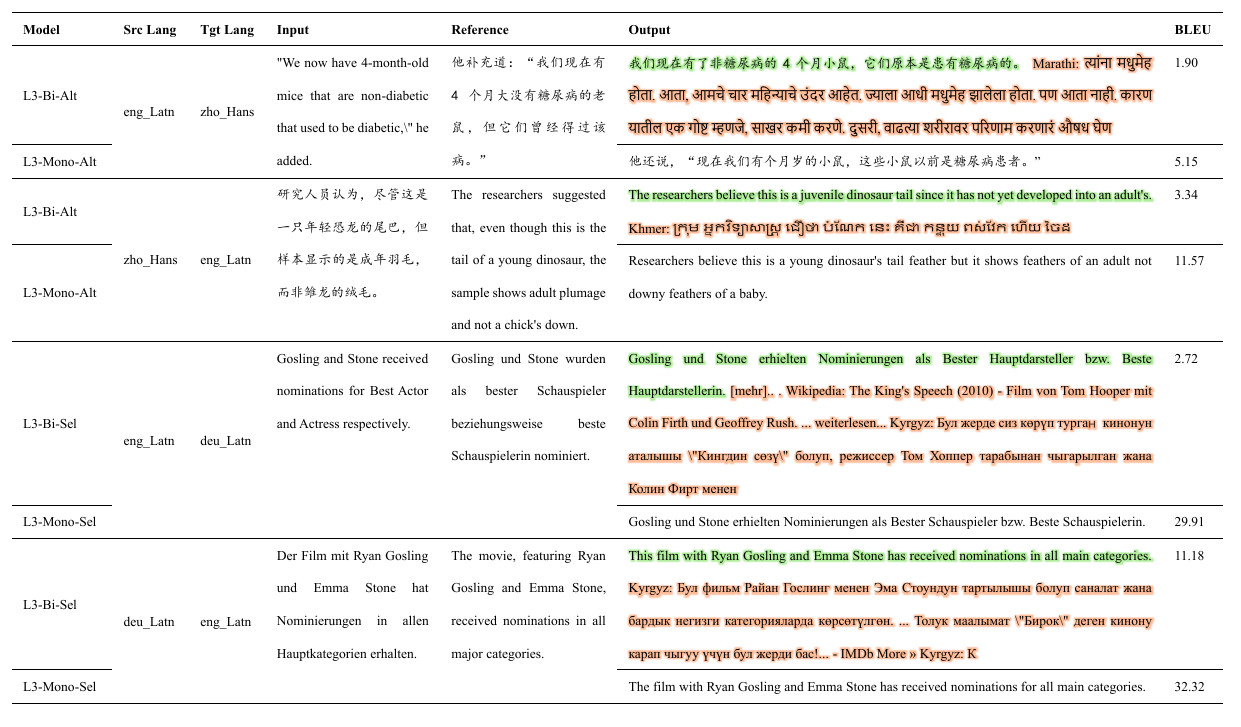} 
\caption{Examples of language mixing in bilingual CPT (L3-Bi-) compared to monolingual CPT (L3-Mono-).}
\label{fig:language_mixing_examples}
\end{figure}

This section supplements the main text with examples of language mixing in bilingual CPT (L3-Bi-), where translations contain unintended multilingual fragments. 
For comparison, outputs from monolingual CPT (L3-Mono-) are provided, showing cleaner, target-language-only results. Individual BLEU scores are included to quantify quality. 
Language mixing reduces BLEU scores by introducing irrelevant tokens that disrupt n-gram precision, as these fragments fail to match the reference translation’s target-language sequences, lowering overlap, especially for higher-order n-grams like the default 4-grams in \texttt{SacreBLEU}~\citep{post2018call}, where a single irrelevant token disrupts multiple overlapping sequences.
Figure~\ref{fig:language_mixing_examples} illustrates the four examples and the translation generated by monolingual and bilingual CPT models.

\subsection{Prompt Templates}
\label{sec:prompt_templates}
For the SIB-200 classification task, we adopt the following template: 
\begin{quote}
    \texttt{Topic Classification: science/technology, travel, politics, sports, health, entertainment, geography.}  
    
    \texttt{\{examples\}}  
    
    \texttt{The topic of the news "\{text\}" is}
\end{quote}

For the FLORES-200 translation task, we employ the following 3-shot prompt:
\begin{quote}
\texttt{Translate the following sentence from \{src\_lang\} to \{tgt\_lang\}}  

\texttt{\{examples\}}  

\texttt{[\{src\_lang\}]: \{src\_sent\}}  

\texttt{[\{tgt\_lang\}]:}
\end{quote}

\subsection{Data Statistics}
The data statistics presented in Tables \ref{tab:bilingual_data_stats} and \ref{tab:monolingual_data_stats} summarize the bilingual translation and monolingual training datasets used in this study. 
Token counts in the two tables are calculated by splitting text on whitespace, a method chosen for its computational efficiency given the large volume of data.

For code data, we provide raw token counts from The Stack dataset across 32 programming languages in Table \ref{tab:code_data_stats}, totaling 51,253,373,176 tokens. We then downsample this to 49,999,171 tokens as counted by using the GPT-2 tokenizer \citep{radford2019language}, selected for its speed, to match the training dataset setup in Subsection~\ref{subsec:pt_data}.

\begin{table}[htbp]
\centering
\small
\setlength{\tabcolsep}{6pt}
\renewcommand{\arraystretch}{1.2}
\begin{tabular}{l l c c c}
\hline
\textbf{Category} & \textbf{Language Pair} & \textbf{Source Tokens} & \textbf{Target Tokens} & \textbf{Total Tokens} \\
\hline

\multirow{10}{*}{Altruistic} 
    & eng\_Latn-zul\_Latn & 12,672,195 & 9,196,313 & 21,868,509 \\
    & zho\_Hani-zul\_Latn & 341,665 & 208,653 & 550,318 \\
    & ceb\_Latn-zul\_Latn & 190,637 & 94,910 & 285,547 \\
    & zho\_Hani-ceb\_Latn & 696,789 & 863,637 & 1,560,426 \\
    & eng\_Latn-mar\_Deva & 7,736,633 & 7,248,634 & 14,985,267 \\
    & zho\_Hani-mar\_Deva & 2,244,545 & 1,825,067 & 4,069,612 \\
    & ceb\_Latn-mar\_Deva & 835,219 & 634,881 & 1,470,100 \\
    & ceb\_Latn-eng\_Latn & 12,355,815 & 11,719,494 & 24,075,309 \\
    & zho\_Hani-khm\_Khmr & 1,157,707 & 577,403 & 1,735,110 \\
    & eng\_Latn-khm\_Khmr & 11,364,386 & 10,147,868 & 21,512,254 \\
\cline{2-5}
    & \textbf{Total} & \textbf{49,595,591} & \textbf{42,516,860} & \textbf{92,112,452} \\
\hline

\multirow{8}{*}{Selfish} 
    & bel\_Cyrl-deu\_Latn & 27,012,850 & 18,085,602 & 45,098,452 \\
    & bel\_Cyrl-eng\_Latn & 1,598,358 & 1,920,079 & 3,518,437 \\
    & deu\_Latn-mri\_Latn & 1,682,621 & 2,250,042 & 3,932,663 \\
    & eng\_Latn-mri\_Latn & 717,914 & 913,809 & 1,631,723 \\
    & deu\_Latn-kir\_Cyrl & 1,682,749 & 1,583,623 & 3,266,372 \\
    & eng\_Latn-kir\_Cyrl & 2,262,374 & 1,515,087 & 3,777,462 \\
    & deu\_Latn-nya\_Latn & 1,155,433 & 1,077,300 & 2,232,733 \\
    & eng\_Latn-nya\_Latn & 19,714,307 & 16,830,192 & 36,544,499 \\
\cline{2-5}
    & \textbf{Total} & \textbf{55,826,606} & \textbf{44,175,734} & \textbf{100,002,341} \\
\hline

\multirow{4}{*}{Stagnant} 
    & eng\_Latn-tha\_Thai & 5,619,794 & 18,138,086 & 23,757,879 \\
    & eng\_Latn-yor\_Latn & 14,334,000 & 16,887,000 & 31,221,000 \\
    & eng\_Latn-sna\_Latn & 9,813,703 & 7,608,164 & 17,421,867 \\
    & eng\_Latn-wol\_Latn & 13,600,133 & 13,636,959 & 27,237,092 \\
\cline{2-5}
    & \textbf{Total} & \textbf{43,367,630} & \textbf{56,270,209} & \textbf{99,637,838} \\
\hline

\end{tabular}
\caption{Bilingual translation data statistics: source, target, and total token counts across language pairs for each language category, with totals for each group.}
\label{tab:bilingual_data_stats}
\end{table}

\begin{table}[htbp]
\centering
\small
\setlength{\tabcolsep}{6pt}
\renewcommand{\arraystretch}{1.2}
\begin{tabular}{l l r}
\hline
\textbf{Category} & \textbf{Language} & \textbf{Total Tokens} \\
\hline
\multirow{6}{*}{Altruistic} 
    & eng\_Latn  & 43,492,709 \\
    & zho\_Hani  & 4,440,706 \\
    & ceb\_Latn  & 14,245,308 \\
    & mar\_Deva  & 9,708,582 \\
    & zul\_Latn  & 9,499,876 \\
    & khm\_Khmr  & 10,725,271 \\
\cline{2-3}
    & \textbf{Total} & \textbf{92,112,452} \\
\hline
\multirow{6}{*}{Selfish} 
    & eng\_Latn  & 24,614,674 \\
    & deu\_Latn  & 22,606,405 \\
    & bel\_Cyrl  & 28,611,208 \\
    & mri\_Latn  & 3,163,851 \\
    & kir\_Cyrl  & 3,098,710 \\
    & nya\_Latn  & 17,907,492 \\
\cline{2-3}
    & \textbf{Total} & \textbf{100,002,341} \\
\hline
\multirow{5}{*}{Stagnant} 
    & eng\_Latn  & 43,367,629 \\
    & tha\_Thai  & 18,138,086 \\
    & yor\_Latn  & 16,887,000 \\
    & sna\_Latn  & 7,608,164 \\
    & wol\_Latn  & 554,809 \\
\cline{2-3}
    & \textbf{Total} & \textbf{86,555,688} \\
\hline
\end{tabular}
\caption{Monolingual training data statistics: total token counts for each language across the three language categories.}
\label{tab:monolingual_data_stats}
\end{table}

\begin{table}[htbp]
\centering
\small
\setlength{\tabcolsep}{6pt}
\renewcommand{\arraystretch}{1.2}
\begin{tabular}{l r}
\hline
\textbf{Language} & \textbf{Total Tokens} \\
\hline
assembly & 331,667,471 \\
c & 8,741,971,474 \\
cpp & 7,816,404,624 \\
c-sharp & 2,378,224,612 \\
clojure & 82,101,240 \\
common-lisp & 392,951,006 \\
dart & 596,729,087 \\
erlang & 145,648,910 \\
f-sharp & 67,025,280 \\
fortran & 442,165,240 \\
glsl & 116,320,040 \\
go & 3,566,871,370 \\
haskell & 401,113,392 \\
java & 3,659,465,643 \\
javascript & 3,027,933,059 \\
julia & 221,192,206 \\
kotlin & 851,638,489 \\
llvm & 383,439,623 \\
markdown & 1,795,961,949 \\
pascal & 424,339,418 \\
perl & 473,210,127 \\
php & 2,315,544,678 \\
powershell & 74,390,317 \\
python & 5,199,071,526 \\
r & 49,449,207 \\
ruby & 1,107,302,714 \\
rust & 1,572,906,932 \\
scala & 568,062,821 \\
shell & 510,858,653 \\
solidity & 151,560,961 \\
sql & 1,179,866,764 \\
typescript & 2,607,984,343 \\
\cline{1-2}
\textbf{Total} & \textbf{51,253,373,176} \\
\hline
\end{tabular}
\caption{Raw code data statistics from a subset of The Stack dataset processed by \citet{ji2024emma}, showing total token counts for each programming language before downsampling.}
\label{tab:code_data_stats}
\end{table}

\end{document}